\documentclass[preprint,12pt]{elsarticle}

\usepackage{amsmath,amssymb,amsfonts}
\usepackage{algorithmic}
\usepackage{graphicx}
\usepackage{textcomp}
\usepackage{booktabs}
\usepackage{colortbl}
\usepackage{tabularx}
\usepackage{adjustbox}
\usepackage{multirow}
\usepackage{tabularx}
\usepackage{caption}
\usepackage{url}            
\usepackage{longtable}
\usepackage{lipsum}
\usepackage{rotating}

\usepackage{subcaption}
\def\BibTeX{{\rm B\kern-.05em{\sc i\kern-.025em b}\kern-.08em
    T\kern-.1667em\lower.7ex\hbox{E}\kern-.125emX}}

\usepackage{amssymb}

\begin{document}

\begin{frontmatter}




\title{Explainable Predictive Modeling for Limited Spectral Data}
\author[inst1]{Frantishek Akulich} \author[inst1]{Hadis Anahideh}
\author[inst1]{Manaf Sheyyab} \author[inst1]{Dhananjay Ambre}

\affiliation[inst1]{organization={University of Illinois at Chicago, Mechanical and Industrial Engineering Department One},
            addressline={842 W Taylor St}, 
            city={Chicago},
            postcode={60607}, 
            state={Illinois},
            country={US}}

\begin{abstract}
Feature selection of high-dimensional labeled data with limited observations is critical for making powerful predictive modeling accessible, scalable, and interpretable for domain experts. Spectroscopy data, which records the interaction between matter and electromagnetic radiation, particularly holds a lot of information in a single sample. Since acquiring such high-dimensional data is a complex task, it is crucial to exploit the best analytical tools to extract necessary information. In this paper, we investigate the most commonly used feature selection techniques and introduce applying recent explainable AI techniques to interpret the prediction outcomes of high-dimensional and limited spectral data. Interpretation of the prediction outcome is beneficial for the domain experts as it ensures the transparency and faithfulness of the ML models to the domain knowledge. Due to the instrument resolution limitations, pinpointing important regions of the spectroscopy data creates a pathway to optimize the data collection process through the miniaturization of the spectrometer device. Reducing the device size and power and therefore cost is a requirement for the real-world deployment of such a sensor-to-prediction system as a whole. Furthermore, we consider a wide range of machine learning models that have been proven to be successful for the prediction of the Cetane Number of fuels. We specifically design three different scenarios to ensure that the evaluation of ML models is robust for the real-time practice of the developed methodologies and to uncover the hidden effect of noise sources on the final outcome. The evaluation is performed for both the full model and reduced models using different feature selection techniques on a real dataset. Finally, we propose a correctness metric for the feature selection techniques to assess the conformance of the selected subset of features to the domain expertise. As a result, the Support Vector Regression yields better prediction accuracy and generalization power as it leads to less complex and computationally more efficient than model Neural Network. More importantly, using the reduced subset of features from original data creates a pathway to deploying less complex, scalable, and explainable prediction models. 
\end{abstract}




\end{frontmatter}



\section{Introduction}




Nowadays, Machine Learning (ML) is widely used in a variety of fields.
ML is broadly designed to aid decision-making processes within organizations. ML refers to a group of computational and statistical techniques that maps a set of input characteristics to a response, with a priori unknown relationships, such that the estimated mapping (model) can be utilized for response prediction of unseen future observations.

There are a group of ML models with explicit functional form, either linear (e.g., Linear Regression \cite{pearson1898mathematical}) or non-linear (e.g., $K$-Nearest Neighbor \cite{altman1992introduction}).
Several problems can be addressed by modeling linear relationships between a response value and a set of predictor variables that a human observing the process easily understands. In contrast, others require a more sophisticated mapping, which gave birth to various non-linear ML modeling techniques. On the other hand, more advanced ML modeling techniques have recently become prevalent, with no explicit functional form associated with the modeling. Instead, the mapping follows an architecture connecting input characteristics to output through a large set of nodes with a chain of transformations. Although such network-based models yield black-box complex functional form, they successfully capture the non-linearity and specific properties of the underlying unknown function. In reality, the nature of the underlying mapping is not known in advance.




Presently, machine learning has been successfully implemented to solve a range of problems from finance, engineering, medicine to applied sciences. In some applications, these models achieve better performance than humans, such as object recognition, detection, and tracking \cite{howard2019searching, krizhevsky2012imagenet}, beating world Chess and Go champions \cite{silver2016mastering, silver2017mastering}, solving protein folding problems \cite{senior2020improved}, etc. Unlike a human, however, a trained black-box model cannot provide reasons for a specific decision or choice, even though the results are often accurate. Therefore, these models lack a very important component of human behavior -- explainability.




Predictive models in practice are often inaccessible to the domain researchers and practitioners since they are typically designed by vendors and thus exclusive. This lack of transparency makes it difficult to understand the prediction outcome and has also led practitioners to reject these products due to a lack of trust.  
Hence, the interpretability of ML models or, in general, AI tools for decision-making processes is inevitable. Interpretability can be defined as a degree to which a human can understand the cause of a decision from an ML model \cite{miller2019explanation}.
System transparency aids better human decisions at critical steps. More importantly, having an interpretable AI system allows incorporating human intelligence to guide the rest of the automated process, making the system more scalable and efficient. This alleviates the pressure to build a perfect automated and black-box AI system and instead focus on constructing a powerful hybrid decision-making process. André Meyer-Vitali et al. \cite{meyer2019hybrid} argue that designing a strategy that uses human judgment and interaction with the system improves the performance of either manual or fully automated systems.

To interpret the ML decision process, there exist two groups of \emph{model-based} and \emph{model-agnostic} techniques. The model-based interpretability is focused on constraining the structure of ML models so that they readily provide useful information about the uncovered relationships. As a result of these constraints, the space of potential models is smaller, which may sacrifice training predictive accuracy \cite{murdoch2019interpretable} to achieve a better generalization performance.
Linear Regression (LR) \cite{pearson1898mathematical}, Logistic Regression \cite{berkson1944application} and Decision Trees \cite{breiman2017classification} are innately interpretable. Therefore, analyzing the interactions between features or between features and response within such models allow the observer to explain the model outcome.
The model-agnostic methods, on the other hand, can be used on any ML model and are usually applied post hoc (post-model) after the model training stage. The widely used preprocessing techniques such as Principal Components Analysis (PCA) \cite{pearson1901liii} and Partial Least Squares (PLS) \cite{wold1966estimation} with its variants, such as Interval PLS \cite{norgaard2000interval}, Forward and Backward Interval PLS \cite{zou2007selection}, Moving Window PLS \cite{jiang2002wavelength} and few others,
are also
independent of the model and can be used before the training stage.
Local and global interpretation methods are two types of model-agnostic interpretation methods \cite{molnar2020interpretable}. Global methods explain the on-average effect of features on the final prediction outcome. Local techniques, however, seek to explain particular predictions. The latter group of techniques is more useful when we do not have access to a lot of data and want to explain the behavior of the model for every single instance of the data.
Most of the local model-agnostic interpretable techniques require a surrogate or a simple proxy model that can be applied to learn a locally faithful approximation of a complex, black-box model based on outputs returned by the black-box model \cite{stiglic2020interpretability}. This approach is also known as knowledge distillation \cite{hinton2015distilling}.
Alternative to local interpretation that helps explain individual prediction, are 
global methods that describe entire model behavior across all predictions. Global methods offer transparency about what is going on inside a model on an abstract level \cite{du2019techniques}. The advantage of Model-agnostic techniques is their flexibility to explain any model, providing consistency in explanation across various prediction methods. Therefore, we can select certain attributes that affect the target outcome using these interpretable methods, which will be summarized later. As mentioned above, some models are interpretable by nature, and some are complex black-box. The biggest advantage of model-agnostic techniques is their ability to explain such complex models. Moreover, this advantage can be extended from complex models to simpler ones as well.


This paper focuses on completing the human-machine loop on the example of a fuel sensor for the next generation of Unmanned Aircraft Systems (UAS) propulsion system. The purpose is to leverage the power of machine learning to accurately predict the ignition properties of fuel through its spectroscopic signature in real-time and at scale. Having access to a limited number of training samples and a large number of attributes results in a more complex problem setting. Hence, the advantage of developing a framework that adopts a subset of attributes for prediction is two-folded, first, making the prediction efficient and at scale in real-time. Secondly, reducing the complexity to avoid overfitting issues on the unobserved noisy test set. Furthermore, the reduced dimensionality of such a tool would enable a more interpretable and accessible model for domain specialists. 

While the data collection is typically peripheral to constructing predictive models, it is essential to scale such models to solve real-world engineering problems. In this paper, we consider the collection of spectroscopy data which is a costly task that requires a robust, sizable sensor. Such sensor, however, can be miniaturized to capture only selected region of spectroscopic signature if, and only if, there is enough information in those regions to differentiate between various types of fuels and their properties. Although in reality such regions are continuous, due to instrument resolution limitations, they are represented by a discrete feature location on wavelength axis.
Therefore, pinpointing these regions creates a pathway to optimize the data collection process, which is a requirement for the real-world deployment of such a sensor-to-prediction system as a whole.

That being said, spectroscopic data of fuels was collected to prepare a framework capable of predicting the fuel ignition quality. Several spectroscopy techniques are used in real-world applications, one of them known as Raman spectroscopy. This is a spectroscopic method that works on the light scattering principle where the incident laser light is scattered by the sample's molecules to produce a unique Raman spectrum. The spectroscopic data can be visualized as a series of data where each feature is recorded as the scattering intensity value occurring at a particular wavelength axis position. This technique is frequently used in chemistry to provide a structural fingerprint by which substrate molecules can be identified. Explicitly, the fundamental Chemical Functional Groups (CFG) present in the sample are the main reason for producing unique chemical fingerprints of the sample under observation. As a unique structural fingerprint is acquired for each sample, fuel ignition quality such as Derived Cetane Number (DCN) can be measured for each sample and associated with its Raman spectral data, thus providing a framework for predicting the DCN of unknown and unseen fuel samples. 

DCN, or just Cetane Number (CN), is one of the main indicators of fuel ignition quality. Similar to spectroscopic and CFG correlations, CN is highly correlated to the quantity and type of functional groups present in the fuel \cite{abdul2016predicting,jameel2018minimalist,dahmen2015novel}. Thus, feature selection is performed in selecting the right functional groups, particularly wavenumbers. This can be validated using the knowledge gained from the physical chemistry analysis, completing the human-in-the-loop approach.



Spectroscopy can suffer from several measurement accuracy issues: poor signal intensity, broad fluorescence baseline interference, and dark current noise. Although various pre-treatment and calibration techniques are often used before the analysis to ensure precise and consistent data readings, both spectroscopy data and measured CN values have an observable level of statistical noise. Notably, these noise levels are correlated to periods of time when data was collected. Bearing that in mind, we divide our dataset into ``old'' and ``new'' sets, depending on when it was collected. The resulting sets are further divided into three separate scenarios: Control (old data only), Mixed (old and new mixed together), and Real-time (old data used for training and new for testing).

While the body of research on spectroscopic analysis is extensive, to the best of our knowledge, there is no single comprehensive work focused on developing both explainable and scalable predictive models for spectroscopic data. Most publications in the field either solely focus on obtaining prediction, for example, applying popular ML methods for octane prediction using infrared spectroscopy \cite{al2019octane}, or the use of common feature elimination techniques \cite{balabin2011variable} to improve the prediction accuracy. The implementation of explainable black-box models is limited to interpreting functional near-infrared spectra data in developmental cognitive neuroscience using simple multi-variate analysis \cite{andreu2021explainable} and using Local Interpretable Model-Agnostic Explanations (LIME) \cite{ribeiro2016should} on optical emission spectroscopy of plasma \cite{wang2021machine}. 

Therefore, the purpose of this paper is to investigate the performance of a wide range of successful predictive models and implement model-based and model-agnostic interpretable techniques to achieve at-scale models for real-time practice. We consider three predictive models, from basic Linear Regression to non-linear Support Vector Machine (SVM) \cite{boser1992training} and network-based Neural Network (NN) \cite{rumelhart1986learning} regressions. Furthermore, we consider PCA, PLS, Ridge Regression \cite{hoerl1970ridge} and Random Forest \cite{ho1995random} for model-based feature selection methods, as well as popular local model-agnostic interpretable techniques, such as LIME and Shapley Additive Explanation (SHAP) \cite{lundberg2017unified}, and Global Surrogate (GS) \cite{vstrumbelj2014explaining} as global model-agnostic method.

Performing a set of comprehensive experiments, we discuss the set of tools for spectroscopy data analysis beyond what has been covered in literature so far and provide informative insights on their challenges in a high-dimensional and limited spectra data setting.

To provide a precise evaluation of feature selection techniques, we propose two metrics: Correctness, which measures selected features' adherence to known chemistry using domain expertise, and Performance, which measures testing error on a predicted CN value. Additionally, we construct a trade-off scale between Correctness and Performance to evaluate the overall accuracy of the above-mentioned techniques in identifying such attributes.


In Section \S~\ref{sec:related}, we analyze the current literature on machine learning, the role of spectroscopic data analysis in chemometrics, and the significance of machine learning in spectra data analysis. We further discuss feature selection and explainable AI techniques research and how these methods help to interpret complex models. Next, we review two major components of our work: predictive modeling and feature selection and interpretation techniques. In Section \S~\ref{sec:predict} we provide background on popular and successful machine learning models, particularly in the chemometrics domain, including SVM and NN. In Section \S~\ref{sec:back} we review feature selection methods which help us identify important spectroscopy features and provide details on interpretation techniques used to explain the behavior of prediction models. In Section \S~\ref{sec:tech} we focus our attention on the practical implementation of these methods and include details about the deployment challenges in the real-time setting. Finally, in Section \S~\ref{sec:results} we provide experimental results and discuss our findings and summarize the work done in Section \S~\ref{sec:conclusion}.

\section{Related Works}\label{sec:related}




The history of predictive modeling dates back to Legendre (1805) \cite{legendre1805memoire} and Gauss (1809) \cite{gauss1809theoria}, who used least-squares linear regression to predict planetary movement. With the advancements of computational hardware in recent times, training highly complex models including Artificial Neural Networks consist of neurons has became prevalent. Artificial neurons were originally proposed in 1943 by McCulloch and Pitts \cite{mcculloch1943logical} and represented a model of the computational unit with multiple inputs, and a binary output . It was later in 1958 when Rosenblatt proposed a full network of neurons, called perceptrons \cite{rosenblat1958perceptron}. Although the field of artificial intelligence was then stagnated for decades, with the official introduction of backpropagation by Rumelhart, Hinton, and Williams in 1986 \cite{rumelhart1986learning}, the interest in neural networks as problem-solving algorithms has exploded. In the meantime, many other learning methods were developed, including Support Vector Machines, developed at AT\&T Bell Laboratories by Vladimir Vapnik and his colleagues in 1963 \cite{vapnik1963pattern} and modified in 1992 \cite{boser1992training}. Support Vector Machines can construct a set of hyperplanes in multi-dimensional space that efficiently separate observations into separate categories, making it one of the most robust models with strong generalization performance. 

Spectroscopic techniques have been widely used for different purposes in various domains such as petrochemical \cite{giwa2015prediction,garcia2019cetane},
medical, pharmaceutical, and biological \cite{blanco1998near,plugge1992use,roggo2007review}, food and agricultural \cite{teye2013review,van1992fourier,buning2003analysis,jahani2020novel}, engineering \cite{pandey2022explainable} and material and geologic \cite{di2020enhanced,shen2021automated} analysis to monitor reactions and conditions of a final product. 

There is an extensive literature in the petrochemical industry and the chemometrics discipline that use spectroscopic data to predict fuel Cetane or Octane Number.
The data is usually collected using the following techniques: Gas Chromatogramy-Mass Spectroscopy \cite{al2019octane,yang2002neural}, Fourier-transform Infrared (FTIR) Spectroscopy \cite{brudzewski2006gasoline,rocabruno2015artificial,barra2020predicting,spiegelman1998theoretical}, Quantitative Structure Property Relationships analysis \cite{kessler2017artificial}, Fatty Acid Methyl Esters (FAME) composition analysis \cite{piloto2013prediction}, Nuclear Magnetic Resonance (NMR) Spectroscopy \cite{abdul2016predicting}, as well as Near Infrared (NIR) \cite{jahani2020novel,balabin2011variable,xiaobo2010variables,cramer2008automated,jouan1995comparison}, Mid-Infrared (MIR) and Raman Spectroscopy \cite{cramer2008automated,li2020determination,spiegelman1998theoretical}. 

Summarizing the existing body of work on spectroscopy analysis (Table \ref{tab:RelatedWork}), it is worth mentioning that the majority of studies on fuel spectroscopy do not collect their own data or have a limited dataset size, as the spectroscopy data collection process is complicated, time-consuming, and expensive. Our project focuses on using Raman spectroscopy since it allows collecting fuel spectroscopy without major preparation or damage to the sample, further aiding the goal of real-world deployment. 
While the majority of the literature cover fuel ignition qualities prediction, the choice of the predictive model is often limited to either linear or non-linear methods, and rarely are both compared. Additionally, fewer topics are dedicated to extracting or explaining (see Interpretability column in Table~\ref{tab:RelatedWork}), spectroscopic features. To this end, they mainly applied model-based feature selection techniques to identify and remove noisy features in order to improve prediction accuracy and computational efficiency \cite{sennott2013artificial, cramer2008automated, jouan1995comparison, li2020machine, al2019octane, wang2019estimating, spiegelman1998theoretical, balabin2011variable, xiaobo2010variables, balabin2011neural}. In another body of work, popular feature selection methods such as PCA \cite{wold1987principal}, and PLS \cite{sjostrom1983multivariate} were employed to discover the correlation between decomposed fuel spectra and fuel sample clustering results \cite{brudzewski2006gasoline}, help isolate certain chemical groups responsible for the deviation in predicted values \cite{jahani2020novel,al2019octane}, and correlate certain spectra regions of pharmaceutical tablets to the concentration of antiviral drug \cite{jouan1995comparison}. 

\begin{table*}[htbp]
  \centering
  \tiny
    \begin{tabular}{|c|c|c|c|c|c|c|c|}
    \toprule
    \textbf{Work} & \textbf{Collected Data} & \textbf{Linear Models} & \textbf{Non-linear Models} & \textbf{Model-based FS} & \textbf{Model-agnostic FS} & \textbf{Scalability} & \textbf{Interpretability} \\
    \midrule
    \ \cite{kessler2017artificial} & -     & -     & +     & -     & -     & -     & - \\
    \ \cite{sennott2013artificial} & -     & -     & +     & -     & -     & +     & - \\
    \ \cite{rocabruno2015artificial} & -     & -     & +     & -     & -     & -     & - \\
    \ \cite{cramer2008automated} & +     & +     & -     & +     & -     & +     & - \\
    \ \cite{garcia2019cetane} & +     & +     & -     & -     & -     & -     & - \\
    \ \cite{noack2013combined} & +     & +     & -     & +     & -     & -     & - \\
    \ \cite{jouan1995comparison} & +     & +     & -     & +     & -     & +     & + \\
    \ \cite{li2020determination} & +     & +     & +     & -     & -     & -     & - \\
    \ \cite{alves2012determination} & +     & +     & +     & -     & -     & -     & - \\
    \ \cite{mendes2012determination} & +     & +     & -     & +     & -     & -     & - \\
    \ \cite{brudzewski2006gasoline} & +     & -     & +     & +     & -     & -     & + \\
    \ \cite{li2020machine} & -     & +     & +     & +     & -     & +     & - \\
    \ \cite{yang2002neural} & +     & +     & +     & -     & -     & -     & - \\
    \ \cite{jahani2020novel} & +     & +     & -     & +     & -     & -     & + \\
    \ \cite{al2019octane} & -     & +     & +     & +     & -     & +     & + \\
    \ \cite{wang2019estimating} & +     & +     & -     & +     & -     & +     & - \\
    \ \cite{barra2020predicting} & +     & +     & -     & -     & -     & -     & - \\
    \ \cite{abdul2018predicting} & +     & +     & +     & -     & -     & -     & - \\
    \ \cite{giwa2015prediction} & -     & -     & +     & -     & -     & -     & - \\
    \ \cite{piloto2013prediction} & -     & +     & +     & -     & -     & -     & - \\
    \ \cite{spiegelman1998theoretical} & +     & +     & -     & +     & -     & +     & - \\
    \ \cite{balabin2011variable} & -     & +     & +     & +     & -     & +     & - \\
    \ \cite{xiaobo2010variables} & -     & +     & +     & +     & -     & +     & - \\
    \ \cite{balabin2011support} & +     & +     & +     & -     & -     & -     & - \\
    \ \cite{andreu2021explainable} & -     & +     & -     & +     & -     & -     & + \\
    \ \cite{wang2021interpreting} & +     & -     & +     & -     & -     & -     & + \\
    \ \cite{wang2021machine} & -     & -     & +     & -     & +     & -     & + \\
    \ \cite{balabin2011neural} & +     & +     & +     & +     & -     & +     & - \\
    \ \cite{cunha2017predicting} & +     & +     & +     & -     & -     & -     & - \\
    \ \cite{abdul2016predicting} & +     & +     & -     & -     & -     & -     & - \\
    \ This Work & +     & +     & +     & +     & +     & +     & + \\
    \bottomrule
    \end{tabular}%
\caption{Summary of related work in comparison to this work}
\label{tab:RelatedWork}%
\end{table*}%

As previously discussed,  few works focus on explaining learning and predictive modeling using spectra data. In \cite{andreu2021explainable} authors applied linear multi-variate analysis to interpret development cognitive neuroscience spectroscopy data. Direct visualization of gradient-weighted class activation mapping of Convolutional Neural Network was developed in \cite{wang2021interpreting} to interpret detection of volatile organic compounds. 

Recently, model-agnostic methods have attracted a lot of attention for feature evaluation, such as
Shapley Additive Explanation (SHAP) \cite{lundberg2017unified} and LIME \cite{ribeiro2016should}.
Explainable AI techniques in general have been widely used to explain predictions in financial and chemical time-series data \cite{ron2019interpreting,rios2020explaining,saluja2021towards,thrun2021explainable} vibrational-based Structural Health Monitoring signals \cite{pandey2022explainable}, hyperspectral imaging \cite{singh2021estimation} and electrocardiogram data \cite{taniguchi2021explainable}. However, to the best of our knowledge, only one recent work focused on using the model-agnostic method (LIME) to explain the non-linear predictions of spectroscopy data to characterize plasma solution conductivity \cite{wang2021machine}.

After investigating the existing research, we develop an efficient, scalable framework that ensures prediction accuracy and transparency together. This paper simultaneously investigates the performance of the most successful ML models in the literature, including SVM and NN, 
and covers a comprehensive examination of both model-based and model-agnostic explainable AI techniques for spectroscopy data analysis. The goal is to derive the most accurate, scalable prediction model. More importantly, we aim to ensure the interpretability and transparency of the prediction outcome to the human expert, supporting the fact that predictions are grounded in domain science and, therefore, can be fully trusted to make further decisions. We also focus on modeling raw, unscaled, noisy data to ensure it can be deployed fast in the real world without any major preprocessing.

\section{Predictive Modeling}\label{sec:predict}

Machine Learning is referred to statistical tools encoded in a machine to make predictions about future observations based on historical data. There are many ways to categorize ML methods; supervised and unsupervised, linear and non-linear, etc. In this paper, we group them into two categories: interpretable and non-interpretable (black-box), with Linear Regression \cite{pearson1898mathematical} being an example of the former one, and Support Vector Machines \cite{boser1992training}, and Neural Networks \cite{rumelhart1986learning} being an example of the latter. As discussed in \S~\ref{sec:related}, the two popular methods in chemometrics are PLS Regression \cite{wold1966estimation}, and PCA Regression \cite{pearson1901liii},  as they can first provide insight into important features in high-dimensional data by decomposing it into latent structures using PCA and PLS, and build interpretable linear regression models on top for prediction.

There is a long tradition of employing linear models in chemometrics. As more recent examples, Jameel et al. used multiple LR and nuclear magnetic resonance spectroscopy to predict fuel ignition quality \cite{abdul2016predicting}. Barra et al. used PLS regression with FTIR spectroscopy to predict cetane number in diesel fuels \cite{barra2020predicting}. Balabin et al. also compared the performance of PLS and PCA regression models with NN while analyzing biodiesel properties using near-infrared (NIR) spectroscopy \cite{balabin2011neural}. However, the main drawback of any linear model is its inability to capture complex non-linear relationships within the data, especially when the access to the training data is limited. Therefore, the use of Support Vector Regression (SVR) and deep learning models has been particularly prominent in chemometrics. Using SVR, Kiefer et al. achieved superior results on Raman spectroscopy of algal production of complex polysaccharides over LR \cite{noack2013combined}, while Alves et al. noted that SVR outperformed PLSR for NIR spectra analysis \cite{alves2012determination}. Balabin et al. has also explored the deployment of NN and SVR for analytical chemistry and concluded that not only SVR outperforms PCA and PLS regression methods \cite{balabin2011neural}, but that SVR also performs similarly to NN, with SVR having the advantage in producing a more generalized model capable of efficiently dealing with non-linear relationships \cite{balabin2011support}. Similarly, NN is also capable of capturing unique spatial features and have been shown to perform well on spectroscopy data analysis \cite{giwa2015prediction,brudzewski2006gasoline,al2019octane,yang2002neural}, given a wealthy amount of data. In our work, we investigate the performance of LR, SVR, and NN models to process high-dimensional spectra data for prediction and explainability effort.


\subsection{Support Vector Regression}

Support Vector Machine is a supervised learning model grounded in Vapnik–Chervonenkis computation learning theory \cite{vapnik1999nature}, which explains the learning process from a statistical point of view, ensuring high generalization ability on unseen data. SVM solves both classification and regression problems by identifying an optimal separating hyperplane with maximum margin to the training observations, formulated as a convex optimization problem. In a regression setting, the optimal hyperplane is the decision surface that best approximates the continuous-valued function \cite{awad2015support}. The goal is to first arrive at minimized convex loss function that produces an error in predicted values at most equal to the specified margin, called the maximum error $\varepsilon$ (epsilon). At the same time, the decision surface must stay as flat as possible while containing most of the training samples \cite{smola2004tutorial}. An important property of SVR is its ability to map input vectors to a high-dimensional feature space where a non-linear decision surface can be constructed that fits the data within a threshold of values within a specified margin. Since the data can often not be separable in initial finite-dimensional space, mapping it into a much higher-dimensional space, aka kernel space, makes the separation easier. 

The optimization, which has a unique solution, is further solved, and since not all points are going to fall within the margins, slack variables $\xi_{n}$ and $\xi_{n}^{*}$, which denote deviation from the margin, are introduced to deal with otherwise infeasible constraints. The constant $C$ is then introduced to impose a penalty on observations that lie outside the margin to prevent overfitting and determine the trade-off between the flatness and amount of deviation that can be tolerated. Ultimately, the decision surface is confined using support vectors, which are the most influential instances that lie outside the tube boundaries and affect its shape. In order to minimize the computational complexity of the described optimization problem, it is solved using Lagrange dual formulation. Mathematically speaking, given the set of observations \textbf{X} where each $X_{i}\in \mathbb{R}^{M}$, for $i=1,\dots, N$ and where $N$ is the number of samples and $M$ is the number of features (dimensions), with $y_{i}$ being the predicted value, we can express the optimization problem as:

\begin{align}\label{eq:svr}
\vspace{-3mm}
&\min \frac{1}{2}||\mathbf{w}||^{2}+C\sum_{i=1}^{N}\xi_{i}+\xi_{n}^{*} \\
\nonumber &\mbox{ s.t. }\\
&y_{i}-\mathbf{w}^{T}\varphi(X_{i})\leq\varepsilon+\xi_{i}^{*}  \quad  i=1,\dots,N\\
&\mathbf{w}^{T}\varphi(X_{i})-y_{i}\leq\varepsilon+\xi_{i},  \quad  i=1,\dots,N\\
&\xi_{n},\xi_{n}^{*}\leq0, \quad  i=1,\dots,N
\end{align}

Where $\textbf{w}$ is the weight vector of the separating hyperplane and $\varphi(\cdot)$ is a transformation function, i.e. kernel, that maps vector $\textbf{X}$ to a high-dimensional space, 
that computes inner products of the input vectors. Using kernel, or kernel trick, pairwise similarity comparisons between training data observations are used instead transforming data to avoid extremely high number of combinations.
Two popular kernel functions include: Linear dot product $\varphi(X_{i},X_{j})=\langle X_{i}\,X_{j}\rangle$ and Polynomial $\varphi(X_{i},X_{j})=(\gamma\langle X_{i},X_{j}\rangle+Coef0)^{d}$, for $d>0$ where $\gamma$ is a scale factor that defines how a support vector shapes the decision surface, and $Coef0$ is an independent term used to overcome dot product computation issues for high-dimensional data. The parameters are further covered in Section \S~\ref{sec:modeling}.

\subsection{Neural Networks}

NN is a computing system that is commonly used for supervised learning. It is represented by a network of artificial neurons or nodes connected by links, where each link has an associated randomly initialized weight and activation level. Each node has an input function (typically summing over-weighted inputs), an activation function, and an output. The weights are updated through a forward and backward propagation until they converge to optimal estimates.

Through the forward approach, the input function of each unit is passed through the activation function, typically a non-linear function, and transformed to a new value that would be passed to the nodes in the subsequent layer. This process is known as \emph{forward propagation}, during which a network learns and creates its own features. Mathematically, the operations in input layer are shown in Equation \ref{eq:Feed1}, where $a^{1}$ is input layer activation, expressed as function $g$ of weights \textbf{w} and training data \textbf{X}. Working forward through the network, the input function of each unit is applied to the weighted sum of the activation on the links feeding into that node. Forward propagation ends at the final (output) layer $L$, which produces a value based on function $h_{w}(X)$ of all previous layer transformations (Equation \ref{eq:Feed2}). Lastly, the total prediction error is calculated using problem-specific loss (or cost) function, e.g. mean squared error (Equation \ref{eq:MSE}), cross-entropy (Equation \ref{eq:CrossEntropy}), so that the gradients which are used to update the weights in next step can be derived. 

\begin{equation}
    \mathbf{a^{1}}= g(\mathbf{w^{1}X})
\label{eq:Feed1}
\end{equation}

\begin{equation}
    h_{w}(X) = \mathbf{a^{L}}= g(\mathbf{w^{L}a^{L-1}})
\label{eq:Feed2}
\end{equation}

\begin{equation}
    J(\mathbf{w})=(1-y)(log(1-\hat{y})+y log(\hat{y})
\label{eq:CrossEntropy}
\end{equation}

Each node $j$ in layer $l$ is ``responsible'' for fraction of the error $\delta_{j}^{l}$ in output nodes it is connected to. Hence, through the backward approach, the error associated with each unit from the preceding layers are back-calculated following Equations \ref{eq:Backprop1}- \ref{eq:Backprop2} and the contributing weights are adjusted, accordingly. This process is known as \emph{backpropagation}, during which partial derivatives of error measurement are calculated to track gradient descent, $\boldsymbol{\Delta}^{l}$, Equation \ref{eq:Gradient} that minimizes cost function until convergence is reached.

\begin{equation}
    \boldsymbol{\delta}^{L}=\textbf{a}^{L}-\textbf{y}
\label{eq:Backprop1}
\end{equation}

\begin{equation}
    \boldsymbol{\delta}^{l}=(\textbf{w}^{l})^{T} \boldsymbol{\delta}^{l+1}(\textbf{a}^{l}*(1-\textbf{a}^{l}))
\label{eq:Backprop2}
\end{equation}

\begin{equation} 
    \boldsymbol{\Delta}^{l} = \boldsymbol{\Delta}^{l}+ \boldsymbol{\delta}^{l+1}\textbf{a}^{(l)^T}
\label{eq:Gradient}
\end{equation}

The choice of network hyperparameters determines the network architecture and how the network is trained and, therefore, is crucially important to obtain a high-performing model. The most common methods of hyperparameter selection are Grid Search, Random Search, and Bayesian Optimization. In this paper, we will adopt Bayesian Optimization to identify the optimal NN architecture. There has been several activation functions proposed in the literature including the most common ones such as Sigmoid and Relu.

\section{Feature Selection}\label{sec:back}

\subsection{Model-based Methods}

\subsubsection{Principal Component Analysis}

PCA is an unsupervised dimension reduction method that transforms data to a new coordinate system with a reduced set of variables that retains most of the information from the original space. The transformed space can be imagined as $P$-dimensional ellipsoid, where longest axis of an ellipsoid represent direction of maximum variance within the data for $P<M$, where 
$M$ is the number of original features (dimensions). Given $N$ number of samples and an $N*M$ matrix of data $\mathbf{X}$, where ${X}_{i}\in \mathbb{R}^{M}$, for $i=1,\dots, N$ and $j=1, \dots, M$, the data has to be standardized to be centered around 0 by subtracting the mean value of each variable from individual data points in a given dimension. The key idea behind PCA is to combine original variables into a set of latent vectors $\mathbf{Z}$ in a linear way, where $Z_{p}\in \mathbb{R}^{M}$ for $p=1, \dots, P$ same as $\mathbf{X}$ since it is its linear combination. More formally, the latent vectors can be presented as follows:

\begin{equation}
Z_{p}=\sum_{j=1}^{M}w_{j}X_{j}
\end{equation}

The latent variables are constructed sequentially. The first projection $\mathbf{z}_{1}$ can be written as:

\begin{equation}
Z_{1}=w_{1}X_{1} + \dots + w_{j}X_{j}, 
\label{equation2}
\end{equation}

Where \textbf{w} is the vector of weights constrained so that its sum of squares is equal to 1. 
According to \cite{bro2014principal}, since the matrix \textbf{X} contains variation relevant to the problem, it seems reasonable to have as much as possible of that variation also in \textbf{Z}. Suppose this amount of variation in \textbf{Z} is appreciable. In that case, it can serve as a good summary of the \textbf{X} variables, hence, allowing us to reduce the number of variables used in the original space. The problem is therefore constructed as maximizing the variance of \textbf{Z} with respect to the optimal weights. By substituting (Equation \ref{equation2}) into mathematical notation for returning the argument \textbf{w} of the maximization function and rewriting it in a vectorized format, we obtain the following objective function:

\begin{equation}
    w_{p} = \operatorname*{arg\,max}_{||\mathbf{w}||=1}  (var(\mathbf{Z})) = \operatorname*{arg\,max}_{||\mathbf{w}||=1}  (\mathbf{w^{T}X^{T}Xw})
\end{equation}

Using the result of the covariance matrix, $\mathbf{X^{T}X}$, we find that the optimal \textbf{w} are therefore the eigenvectors of $\mathbf{X^{T}X}$, where the first ordered eigenvector corresponds to $w_{1}$ and so on. Scaled by the variance, which is squared root of corresponding orthogonal eigenvalues, columns of \textbf{w} are therefore called loadings in PCA. As a result, reconstructed back to the original space and applied to fuel spectroscopy, the output of the PCA model contains information on scores of various features that contribute most to the explained variance in recorded absorption values.


\subsubsection{Partial Least Squares}
PLS is a supervised alternative to PCA and is arguably the most widely used technique in the chemometrics domain. Similar to PCA, it identifies a set of features that are linear combinations of the original features. 
Unlike PCA, however, PLS considers the response $\mathbf{y}$ and finds the multidimensional coordination of the feature space $\mathbf{X}$ that contributes to most variance in the 
response $\mathbf{y}$. Essentially, partial least squares seek directions that have high variance and have a high correlation with the response, in contrast to principal components regression which keys only on high variance \cite{hastie2009elements}. 
After standardizing the data to have 0 mean and variance of 1, both $\mathbf{X}$ and $\mathbf{y}$ are decomposed as a product of a common set of orthogonal vectors and a set of specific loadings \cite{abdi2003partial}. The data and response matrices are decomposed as:

\begin{equation}
    \begin{aligned}
    \mathbf{X} = \mathbf{ZP}^{T}
    \\[1pt]
    \mathbf{y} = \mathbf{ZQ}^{T}
    \end{aligned}
\label{eq:PLSdecomp}
\end{equation}

Where $\mathbf{Z}$ is a matrix of latent vectors, and $\mathbf{P, Q}$ are the loading coefficient matrices. A rank regression is then performed to construct a matrix of latent components $\mathbf{Z}$ as linear transformation of $\mathbf{X}$, where $\mathbf{w}$ is a vector of weights:
\begin{equation}
    \mathbf{Z=Xw}
    \label{eq:PLSreg}
\end{equation}

The idea behind PLS is to perform decomposition so that the information from both $\mathbf{X}$ and $\mathbf{y}$ is taken into account. 

The elements of the weight vector $\mathbf{w}$ are defined such that the squared sample covariance between response and the latent components is maximal under the condition that the latent components are mutually uncorrelated \cite{boulesteix2007partial}. Finally we adopt the following objective function to find optimal set of weights for each latent vector $p=1, \dots, P$, that we later use for feature selection:

\begin{equation}
    w_{p} = \operatorname*{arg\,max}_{\mathbf{w}}  (\mathbf{w}^{T}\mathbf{X}^{T}\mathbf{y}\mathbf{y}^{T}\mathbf{Xw})
    \label{eq:PLSobjective}
\end{equation}

\subsubsection{Random Forest}
Random Forest is an ensemble learning method, which constructs multiple de-correlated decision trees trained on different subsets of the data and subsets of selected attributes to reduce the variance in $\mathbf{y}$. Although Random Forest is a prediction tool, it is widely used to 
rank the importance of the variables based on the number of times  
they are used during node splitting. 
For each node $t=1, \dots, K$, where K is total number of nodes within a binary tree $r=1, \dots, T$, where $T$ is number of trees of the random forest, the optimal split in a classification setting is determined by impurity, measuring how well a potential split separates observations that are similar to each other \cite{menze2009comparison}. In regression problems, the measure of impurity ($i(t)$) is variance of the predicted value of observations within each partition.
Therefore, the importance ($w$) of a feature ($j$) is computed as the (normalized) total reduction of variance ($\Delta{i}$) brought by that feature across all trees. Mathematically it can be expressed as:
\begin{equation}
    w_{j}=\sum_{r=1}^{T}\sum_{t=1}^{K}\Delta i_{j}(t,r)
\end{equation}

\subsubsection{Ridge Regression}
The linear regression regularization method is used to reduce model complexity by adding penalties to coefficients of variables in the cost function, such that the Residual Sum of Squares (RSS) loss function for $i=1, \dots, N$ and $j=1, \dots, M$, where $N$ is number of samples and $M$ is number of data dimensions, takes the form:
\begin{equation}
    RSS = \sum_{i=1}^{N}(y_{i} - 	\hat{y_{i}})^2 + \alpha\sum_{j=1}^{M} w^2_j
\end{equation}
The first term in the above equation is the sum of squared error and the second term is the regularization component. $\alpha$ is the penalty parameter used over all weights of features $\mathbf{w}$ to shrink the magnitude of the unimportant ones to ensure that the model does not overfit.

Tuning $\alpha$ hyperparameter controls the strength of the penalty term or essentially the amount of shrinkage, which results in sparse models, easier to analyze than high-dimensional data models with large numbers of parameters.



\subsection{Model-agnostic Methods}\label{sec:agnostic}

\subsubsection{SHAP Method}
SHAP stands for Shapley additives Explanations, which is considered a popular state of the art in explaining black-box machine learning models. SHAP is a technique to calculate the impact of each feature on the prediction outcome using Shapely Values. Shapley values were introduced in the 1950s by Lloyd Shapley \cite{shapley1953value}, who introduced it as a solution concept in cooperative game theory. The main idea is that any model output does not rely only on one single feature but on the entire set of features in the data set. 

Suppose that we have a predictive model, where
the game outcome represents the model prediction, and the players in the game represent the features.
Considering all possible coalition among the players (features) and their effect on the game (model outcome), each player contributes to the team's result. The sum of the contributions for each player from each possible coalition returns the value of the target variable (model outcome) given a particular feature.
As a result, Shapely Values calculates the contribution of each feature to the target value, which is referred to as local marginal contribution or local Shapley Values.
Repeating the same process using combinatorial calculus and retraining the model over all possible combinations of features, 
we can calculate all local Shapley Values for a specific feature. The average absolute value of the local Shapley Values can be used as a measure of feature importance. 

More formally, let $S$ be a subset of features that does not include the feature for which we calculate the importance. Let $M$ be the full set of features. Given a model $g(x)$ trained to predict $f(x)$, the marginal contribution of feature $i$ to the model's prediction and accordingly to the $f(X)$ is:
\begin{equation}
    w_j = \sum_{S \subseteq M \smallsetminus j } \frac{|S|!(|M| - |S| - 1)!}{|M|!} [g(S\cup j) - g(S)],
\end{equation}
Where $S\cup j$ is the subset that includes features in $S$ plus feature $j$ and $S \subseteq M \smallsetminus i$ indicates all sets $S$ that are subsets of the full set of features $M$, excluding feature $j$.

The computation time increases exponentially with the number of features. One solution to keep the computation time manageable is to compute contributions for only a few samples of the possible coalitions. Lately, Lundberg and Lee developed an algorithm for interpreting model predictions \cite{lundberg2017unified}, which uses the Shapely Values to reverse-engineer the output of any predictive algorithm and identify features' contributions. SHAP  approximates the conditional expectations of SHAP values by using a selection of background samples to reduce the computation time. 
By aggregation over multiple background samples, 
SHAP estimates values such that they sum up to the difference between the expected model output on the passed background samples and the current model output $(f(x) - E[f(x)])$. Features that contribute the most to the difference between the expected model output on the passed background samples and the naive case prediction are chosen as important features by SHAP. SHAP method can be used to analyze the prediction for both classification and regression models.

\subsubsection{LIME}
The LIME explanation method was originally proposed by Ribeiro et al. in 2016 \cite{ribeiro2016should}. The key idea of LIME is to locally approximate a black-box model by a simpler glass-box model such as Linear Regression or Random Forest, which is easier to interpret. Such an interpretable model must be locally faithful, meaning it must correspond to how the black-box model behaves in the vicinity of the instance being predicted. LIME works by perturbing any individual data point and generating synthetic data, which gets evaluated by the black-box model and is ultimately used as a training set for the simple model. The variables are perturbed by sampling from a normal distribution and doing the inverse operation of mean-centering and scaling the values according to the means and standard deviations in the original training set. The LIME is capable of explaining any model, and thus it is model-agnostic. The aim of LIME is to minimize a loss function $L(f, g, \pi_{x})$ that can be expressed mathematically in the following way:

\begin{equation}
    \xi(x)=\operatorname*{arg\,min}_g L(f,g,\pi_{x}) + \Omega(g)
\label{eq:LIME}
\end{equation}

Where $f$ is the original model, $g$ is an interpretable model, and $\pi_{x}$ is the similarity kernel that measures the proximity of a new perturbed sample point $z$ to the original data point $x$. Additionally, the $\Omega(f)$ is referred to as a measure of complexity, opposed to interpretability. The Equation \ref{eq:LIME} demonstrates that perturbed samples are generated around $x$ and weighted by $\pi_{x}$ to approximate $L(f, g, \pi_{x})$. Using this approximation LIME can explain local behaviour of the original model $f$ and measure the relative error between the explanation $\xi(x)$ and the original model predictions.

\subsubsection{Global Surrogate}

Similar to LIME, the global surrogate model is used to approximate the predictions of highly non-linear ML models with simpler, interpretable models. Since it is a global method, the surrogate model tries to mimic the function of the entire black-box model to understand its overall behavior. A global surrogate model does not require any information on how the original black-box model works and thus is considered model-agnostic. The process of training a surrogate involves obtaining the predictions of the black-box model on the training dataset $X$. Then, a selected interpretable model is trained on $(X,\hat{Y})$
using black-box predictions as targets. The surrogate model can be any interpretable model, such as Linear Regression, Decision Tree, K-nearest neighbor, or any model that the coefficients could provide insights into the model behavior. The ability of the surrogate to capture the behavior of the black-box model is estimated by computing the error between surrogate and black-box predictions, typically using the r-squared score. A caveat of global surrogate models is that the performance of the underlying black-box model in predicting the actual outcome plays no part in training the interpretable black-box model
\cite{molnar2020interpretable}.

\section{Explainable Prediction for Spectra Data}\label{sec:tech}
\subsection{Domain}
In the Physical Chemistry and Combustion domain, it is well established that chemical functional groups are the building blocks of any hydrocarbon structure such as fuels. A functional group can be defined as any portion of a molecule composed of a group of atoms, which governs both its physical and chemical properties as well as chemical reactivity \cite{rezakazemi2013development}. Practical fuels, such as gasoline, comprise a large number of hydrocarbon molecules, which in turn might contain different functional groups. Knowing the quantity and type of functional groups present in a compound allows researchers to determine its properties. One of such properties is fuel ignition quality, which is an indicator of the combustion speed of the fuel. 
Fuel ignition quality is one of the most important properties that scientists from the applied combustion field have been working for over the years. It is critical to measure ignition quality and correlate it to the fuel's functional groups. 
The functional groups have unique properties resulting in different light absorption values captured during spectroscopy analysis and data collection. Therefore, this paper aims to detect their location on the spectral band using explainable AI tools and feature selection techniques.

In the context of predicting fuel ignition properties, the task of feature selection is to select a subset of wavelengths that improves model interpretability (further called \emph{correctness}) and at-scale prediction performance. 
In a model with higher correctness, selected features must represent locations of chemical functional groups that determine fuel ignition properties, making the overall process transparent to the human operator. Involving a group of domain experts, we choose 120 wavenumbers \cite{george2001infrared} that correspond to chemical bonds in different functional groups within our data and compare them to the outputs of the above-listed feature selection techniques (Section \S~\ref{sec:back}). Since measured wavenumber location can deviate due to instrument noise, we generate ``bins'' that capture the discrete output of attribute selection methods and match it to the expected locations. 

Note that each functional group stretches over certain regions of spectra, which can be identified by multiple peaks in that region. The exact locations of these peaks are unknown, however, using the literature on pure components \cite{george2001infrared} that make up the mixtures, we can try to estimate their locations. Each peak is considered as a feature selected by the expert $S_{Ex}$.

Furthermore, there is a need for miniaturized spectroscopic instruments for high-profile applications in collecting the respective spectra for detailed analysis. Feature selection opens up a pathway towards such miniaturization. By selecting the important subset of wavelengths over the entire range of the spectrum, which affects the prediction performance of the output variable, it is possible to group the features into regions of importance. Once these regions are known, the spectroscopic instrument can be miniaturized by selecting filters such that the data is only collected for the important subset of wavelengths. As the scope of the spectral range has reduced, a smaller instrument could be designed for the same application without trading off the prediction accuracy. Also, another approach could be to use lower resolution instruments which would require smaller components and thus a smaller instrument.

\subsection{Dataset}
The spectroscopy data used in this paper was collected at UIC High-Pressure Shock Tube Laboratory using a Raman spectrometer. The entire dataset includes 245 observations from 49 unique fuel samples collected at various times. The first 145 observations (based on 29 unique mixtures replicated five times each) were collected at a single session over four hours nonstop, where the laser power stayed consistent with minimizing the instrument noise. Note that environmental conditions (e.g., temperature) are exogenous factors that can affect the laser power. The second 100 observations are collected later in a separate session over two hours. This separate data collection allows us to consider noise in the data for our scenario analysis. During the collection of the old dataset's observations, the laser power was measured as 350.5 mW, and for the new dataset, it was measured as 364.2 mW. 

For both collected datasets, the resolution was set to 7.1 ${cm}^{-1}$, with wavelength range between $52.52$ and $3712.89$ ${cm}^{-1}$. The Raman laser power was set at the maximum laser power setting 
to ensure consistent readings for all observations. Each sampled spectra data initially had 2048 features (intensity values at different wavelength locations) and a measured CN value as the response. 
The datasets were pre-processed to filter out highly noisy regions outside of wavelengths range $[181.45, 3200.82]$ ${cm}^{-1}$, resulting in 1562 features. Note that the fingerprint region normally starts at 500 ${cm}^{-1}$, however, in this paper, we aim to investigate the presence of functional groups in the [181.45-500] ${cm}^{-1}$ region. We specifically trimmed [50.52-181.45] ${cm}^{-1}$ since no absorption value is recorded in this region. Similarly, we removed the data in [3200.82, 3712.89] ${cm}^{-1}$ region because no functional groups were present for the considered fuel mixtures.

For simplicity we call the first 145 collected observations as \emph{old} dataset and the 100 observations as \emph{new} dataset.

Three scenarios are constructed accordingly to observe the effect of noise on feature selection and prediction accuracy that is inevitable in the real-time practice of ML for ignition delay. The \emph{Control} scenario includes 145 old observations, and the \emph{Mixed} scenario includes 245 old, and new observations merged, both split into training, testing and validation set 80/10/10. As for the \emph{Real-time} scenario, 145 old data points are split into training and validation sets 80/20, while the new 100 points are used as the testing set. 

Using the preprocessed dataset for each scenario the reduced subset of attributes is identified implementing considered feature selection techniques including \textbf{PCA}, \textbf{PLS}, \textbf{RF}, \textbf{Ridge}, \textbf{SHAP}, \textbf{GS}, and \textbf{LIME}. Each predictive model, that shall be elaborated in Section \S~\ref{sec:modeling}, is then fitted on the reduced subsets for each scenario as well as full set of 1562 features of data. The former is referred to as \emph{Full model} and the latter as \emph{Reduced model}s.

\subsection{Predictive Modeling For Spectra Data}\label{sec:modeling}

To implement ML models for CN prediction there are hyperparameters associated with each considered models discussed in Section \S~\ref{sec:predict} including \textbf{SVR} and \textbf{NN}. Due to the small overall dataset size and to avoid data splitting and model training bias, the data is randomly split 30 different ways using a random seed generator during each model development. The performance of the models is assessed based on the mean ($\mu$) and standard deviation ($\sigma$) across 30 random executions. To choose optimal hyperparameters, we perform Bayesian Optimization (BO) using a predefined range of parameters for SVR (Table \ref{tab:SVRparameters}) and NN (Table \ref{tab:NNparameters}) for both Full and Reduced models trained on various subsets of features for each scenario. We use epsilon-SVR from the libsvm \footnote{\url{https://pypi.org/project/libsvm/}} package for SVR modeling and Keras \footnote{\url{https://keras.io/guides/sequential_model/}} sequential model package to construct our NN.

\begin{table*}[!htbp]
  \centering
  \begin{scriptsize}
    \begin{tabular}{|p{20.335em}|p{19.335em}|}
    \toprule
    \rowcolor[rgb]{ .651,  .651,  .651} \textbf{Parameters} & \textbf{Options} \\
    \midrule
    Kernel & ['poly', 'rbf', 'sigmoid','linear'] \\
    \midrule
    Degree ($d$)& [1 : 4] \\
    \midrule
    Gamma ($\gamma$)& [0.0001 : 1] \\
    \midrule
    Coef0 & [0.01 : 10] \\
    \midrule
    C     & [0.1 : 1000] \\
    \midrule
    Epsilon ($\varepsilon$) & [0.01 : 10] \\
    \bottomrule
    \end{tabular}%
      \end{scriptsize}

\caption{SVR hyperparameters considered for optimization}
\label{tab:SVRparameters}
\end{table*}%

\begin{table*}[!htbp]
  \centering
  \begin{scriptsize}
    \begin{tabular}{|p{15.835em}|p{25.915em}|}
    \toprule
    \rowcolor[rgb]{ .651,  .651,  .651} \textbf{Parameters} & \textbf{Options} \\
    \midrule
    Activation (hidden) & ['relu', 'sigmoid'] \\
    \midrule
    Number of hidden layers & [1 : 10] \\
    \midrule
    Hidden units & min\_value=32, max\_value=8000, step=32 \\
    \midrule
    Activation (output) & ['linear', 'sigmoid'] \\
    \midrule
    Optimizer & ['adam', 'sgd', 'rmsprop'] \\
    \midrule
    Learning rate & [1e-4 : 1.0] \\ 
    \midrule
    Kernel regularization & [0.0001 : 0.01] \\
    \midrule
    Kernel weight initializers & ['random\_normal', 'glorot\_uniform', 'he\_normal’] \\
    \midrule
    Batch size & [32 : 100] \\
    \midrule
    Epochs & [100 : 1000] \\
    \midrule
    Architecture & ['up', 'down', 'up-down', 'down-up', 'random'] \\
    \bottomrule
    \end{tabular}%
      \end{scriptsize}
\caption{Keras Neural Network hyperparameters considered for optimization}
\label{tab:NNparameters}
\end{table*}

For SVR, the regularization hyperparameter $C$ is a free parameter that trades off the influence of higher-order versus lower-order terms in the polynomial. $Gamma (\gamma)$ is a scaling parameter that controls the shape of Support Vector curvature, allowing it to fit the peaks observed in our data. $Epsilon (\varepsilon)$ is a margin term that allows more points to be included in the decision surface without penalizing them during the training. The choice of the kernel in SVM determines the shape of the transformed high-dimensional hyperplane and allows to avoid complex calculations, while the parameters $Degree (d)$ and $Coef0$ are typically used for the polynomial kernel to determine the degree of polynomial fit and adjust the independent term accordingly. 
It is worth mentioning that the (implicit) feature space of a polynomial kernel is equivalent to that of polynomial regression, but without the combinatorial blowup in the number of parameters to be learned \cite{goldberg2008splitsvm}. The algorithm used in epsilon-SVR calculates the outer product of two vectors of features (or a vector with itself), which can be used as an approximation of the polynomial kernel feature space instead of explicitly computing the outer product, which can be extremely inefficient. The resulting kernel space has the same dimensions as original training data, while full third-degree polynomial expansion of 1562 features would result in over 620 million features.

The optimal hyparameters of SVR for both Full and Reduced models are obtained using open source Bayesian Optimization tool \footnote{\url{https://github.com/fmfn/BayesianOptimization}} as: \emph{Kernel='poly'}, \emph{Degree (d)=3}, \emph{Gamma ($\gamma$)=0.7}, \emph{C=0.7}, \emph{Coef0=0.1}, \emph{Epsilon ($\varepsilon$)=0.1}, with maximum number of iterations of the solver fixed to 100,000. This model is therefore referred to as BO-Tuned SVR.

The complexity of the SVR model is defined based on the number of support vectors defining the decision boundary. Table \ref{tab:svr_complexity} illustrates the complexity of the SVR model for each scenario. Since the reduced subsets are created based on a smaller subset of individual wavenumbers to capture the complexity of the spectra data, the model needs a larger number of support vectors to define the decision boundary. For example, in Table \ref{tab:svr_complexity} we observe that 50 support vectors are selected for the Full model setting, and on average, 63 support vectors are necessary for the Reduced setting in a Real-time scenario. 
Moreover, the number of support vectors 
increases from 50 in the Control scenario (145 observations) to 113 vectors in the Mixed scenario (245 observations) for the Full model setting, where the noisy observations were included. This is also reflected in the average runtime shown in Table \ref{tab:finalResults} as training time increases drastically for SVR with the addition of new noisy samples \footnote{Note that the values represented under column ``Time (s)'' is the total computation time of 30 executions.}. For example, the average training time for the BO-Tuned SVR model in Control scenario is 0.17 seconds, compared to 1.4 seconds in the Mixed scenario. 

\begin{table}[!htbp]
  \centering
    \begin{scriptsize}

    \begin{tabular}{|l|c|c|c|c|}
\cmidrule{2-4}    \multicolumn{1}{r|}{} & \multicolumn{3}{c|}{\textbf{\# of Support Vectors}} & \multicolumn{1}{r}{} \\
\cmidrule{2-4}    \multicolumn{1}{r|}{} & \textbf{Control} & \textbf{Mixed} & \textbf{Real-time} & \multicolumn{1}{r}{} \\
    \midrule
    \textbf{Full Model} & 50    & 113   & 50    & \multirow{9}[18]{*}{\begin{sideways}\textbf{BO-Tuned SVR}\end{sideways}} \\
\cmidrule{1-4}    \textbf{Expert} & 52    & 107   & 52    &  \\
\cmidrule{1-4}    \textbf{PCA} & 73    & 126   & 73    &  \\
\cmidrule{1-4}    \textbf{PLS} & 79    & 132   & 79    &  \\
\cmidrule{1-4}    \textbf{RF} & 64    & 122   & 64    &  \\
\cmidrule{1-4}    \textbf{Ridge} & 66    & 122   & 66    &  \\
\cmidrule{1-4}    \textbf{SHAP} & 61    & 121   & 61    &  \\
\cmidrule{1-4}    \textbf{GS} & 63    & 128   & 63    &  \\
\cmidrule{1-4}    \textbf{LIME} & 48    & 119   & 48    &  \\
    \midrule
    \multicolumn{1}{|r}{} & \multicolumn{1}{c}{} & \multicolumn{1}{c}{} & \multicolumn{1}{c}{} &  \\
    \midrule
    \textbf{Full Model} & 33    & 67    & 33    & \multirow{9}[18]{*}{\begin{sideways}\textbf{Fine-Tuned SVR}\end{sideways}} \\
\cmidrule{1-4}    \textbf{Expert} & 28    & 63    & 28    &  \\
\cmidrule{1-4}    \textbf{PCA} & 36    & 77    & 36    &  \\
\cmidrule{1-4}    \textbf{PLS} & 53    & 88    & 53    &  \\
\cmidrule{1-4}    \textbf{RF} & 34    & 75    & 34    &  \\
\cmidrule{1-4}    \textbf{Ridge} & 28    & 60    & 28    &  \\
\cmidrule{1-4}    \textbf{SHAP} & 37    & 68    & 37    &  \\
\cmidrule{1-4}    \textbf{GS} & 33    & 66    & 33    &  \\
\cmidrule{1-4}    \textbf{LIME} & 32    & 70    & 32    &  \\
    \bottomrule
    \end{tabular}%
            \end{scriptsize}

    \caption{SVR Complexity}

  \label{tab:svr_complexity}%

\end{table}%

\begin{figure}
\centering
\subfloat[SVR]{\includegraphics[width=0.49\textwidth]{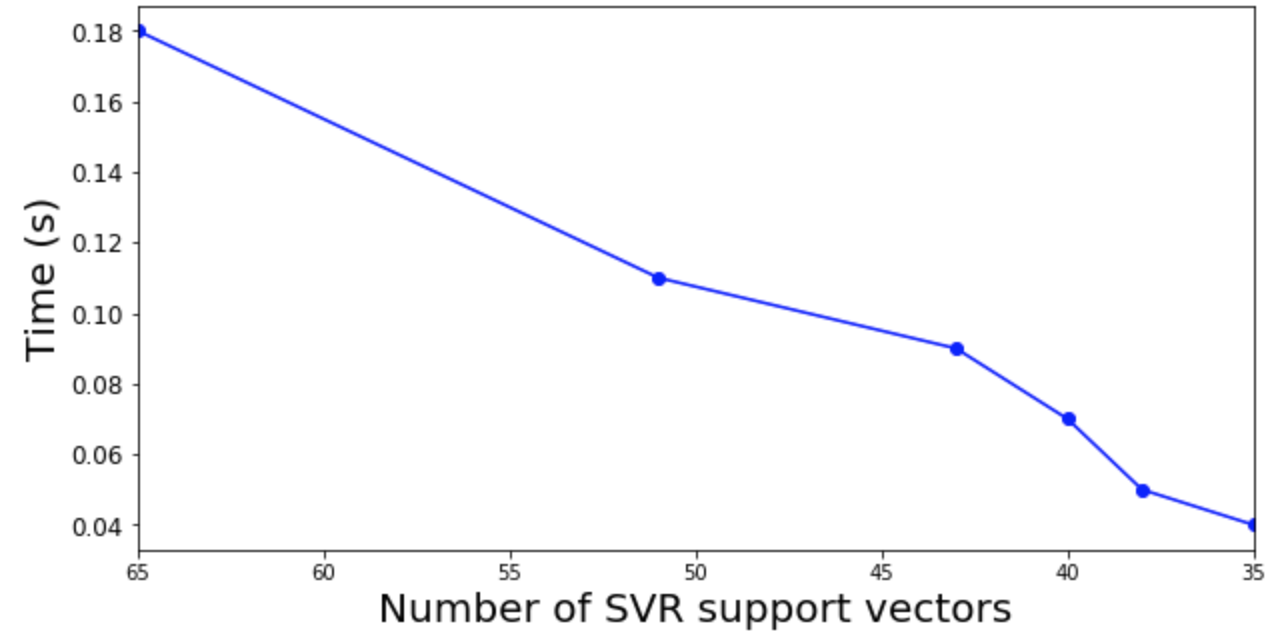}}
\subfloat[NN]{\includegraphics[width=0.49\textwidth]{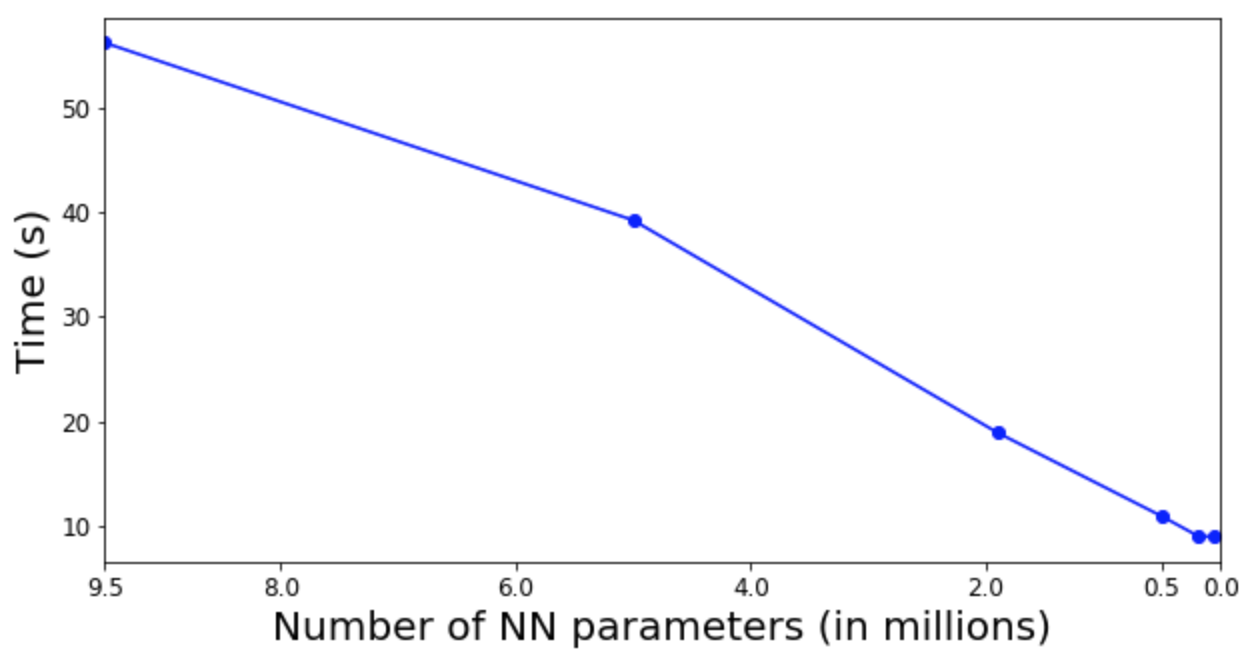}}
    \caption{Average computation time of (a) SVR (b) NN on reduced subsets of data for Real-time scenario with various model complexity levels}
\label{fig:ComputeAandB}
\hspace{\fill}
\vspace{-7mm}
\end{figure}

Since our goal is to ultimately deploy a strong predictive model in real-time, it requires a certain level of generalization to ensure accurate prediction of new, previously unseen observations. The SVR hyparameter $Epsilon (\varepsilon)$ plays a significant role in the generalization power of the model. While BO maximizes exploitation of training data distribution, fine-tuning this parameter allows the construction of decision surfaces both accurate in shape and wide enough to generalize to unseen fuel spectroscopy samples, as will be shown in the Real-time scenario when test data is noisy. Using softer epsilon margin ($Epsilon (\varepsilon)=0.66)$ results in a reduced number of support vectors (33 compared to 50 for Full model in Real-time scenario), which further simplifies the model and makes computation more efficient (Figure \ref{fig:ComputeAandB}a), aiding our scalability effort. More importantly, fine-tuned SVR drastically improves the performance of SVR for our Real-time scenario, in some cases decreasing testing error by the factor of 10.

When constructing Neural Network, the following hyperparameters are considered. \emph{Kernel weight initializer} determines the distribution of weights associated with each layer before the commencement of training and their consequent update through backpropagation. The \emph{number of hidden layers} and \emph{Hidden units} determines the power of the network to perform a linear or non-linear transformation on inputs and guides over model complexity. An \emph{Activation} function in a neural network defines how the weighted sum of the input is transformed into an output from a node or nodes in a layer of the network \cite{brownlee2019choose}. The \emph{Kernel regularization}, also known as weight decay, is aimed at reducing the likelihood of model overfitting by keeping the values of the weights and biases small. \emph{Optimizer} is an algorithm used to adjust model parameters (weights) to maximize a selected loss function (in our case, it's \emph{mean squared error}), while the \emph{Learning rate} determines the rate of adjustment. \emph{Batch size} defines a number of samples to propagate through the network at one time, and \emph{Epochs} is a measure of the number of cycles it takes to train the network with all training data.

We select optimal hyperparameters for Full and Reduced models using KerasTuner \footnote{\url{https://keras.io/api/keras_tuner/tuners/bayesian/}} framework with the objective function aimed at reducing validation Mean Squared Error (MSE). The MSE mathematical formula is provided in Equation \ref{eq:MSE}. The global random seed is set to ensure consistent kernel weight initialization and reproducible results.
We also include an architecture-specific hyperparameter that dictates whether the overall architecture shape expands/shrinks or is produced at random in the optimal setting, 
for each consecutive layer of the network. 
We introduce a set of constraints to limit the exponentially large solution space of hyperparameters and make the optimization computationally stable while exploring versatile architectures. The number of nodes and layers are given the flexibility to be chosen at random or follow the pattern where
the number of nodes is doubled or reduced by a factor of 2 for each subsequent hidden layer and constrained to a maximum of 8,000 nodes. The widening and shrinkage of the network can be both symmetric or asymmetric, with respect to the number of layers before and after the layer with minimum/maximum number of nodes. In an asymmetric case, the number of layers and nodes is selected randomly after picking layers with minimum/maximum nodes.

Given the budget of 1000 trials for BO, the optimal Full model architecture is determined to consist of the input layer, one hidden layer, and a final output layer with \emph{1562-5984-1} nodes. The optimal Reduced model is selected using an asymmetric architecture that
has input, four hidden layers and output layer with \emph{512-1024-2048-4096-512-1} nodes. 
Total of 11,800,000 and 13,170,000 trainable parameters are used in the Full and Reduced models, respectively. The optimal selected \emph{Activation (hidden)} function for the input and hidden layers is \emph{'sigmoid'} for the Full model and is \emph{'relu'} for the Reduced model. Other hyperparameters for both Full and Reduced NN models are selected as: \emph{Batch size=32}, \emph{Learning rate=0.0011} and \emph{Optimizer='Adam'}. 
No significant difference is found using different kernel regularizers and weight initializers. Therefore no regularization was used, and weights were set using default \emph{glorot\_uniform} initializer. Other parameters are fixed to \emph{Epochs=200} and \emph{Activation (output)='linear'}.

When dealing with high-dimensional limited sample data, Network models lead to overfitting and model instability due to highly variant gradients \cite{liu2017deep}. Hence, identifying the optimal Reduced model architecture requires more complexity and includes more trainable parameters than the Full model during the optimization step. Such behavior can be justified by that the network models tend to compensate through the generation of new internal features to capture hidden non-linearity in the Reduced sparse setting. 
Unlike SVR, the number of parameters for NN is fixed and does not increase with the addition of new noisy data. As a result, computational time increases marginally, which can be observed in average runtime in Table \ref{tab:finalResults}. After testing the final model in the Real-time scenario, we observe that the model performs worse than the Full model. Therefore, hyperparameter optimization is biased towards reducing the validation error that follows the training set distribution. Hence, Bayesian optimization does not find the optimal model capable of generalizing on unobserved data. 

\begin{figure}[!htbp]
\centering
\begin{subfigure}{\textwidth}
    \includegraphics[width=1\linewidth]{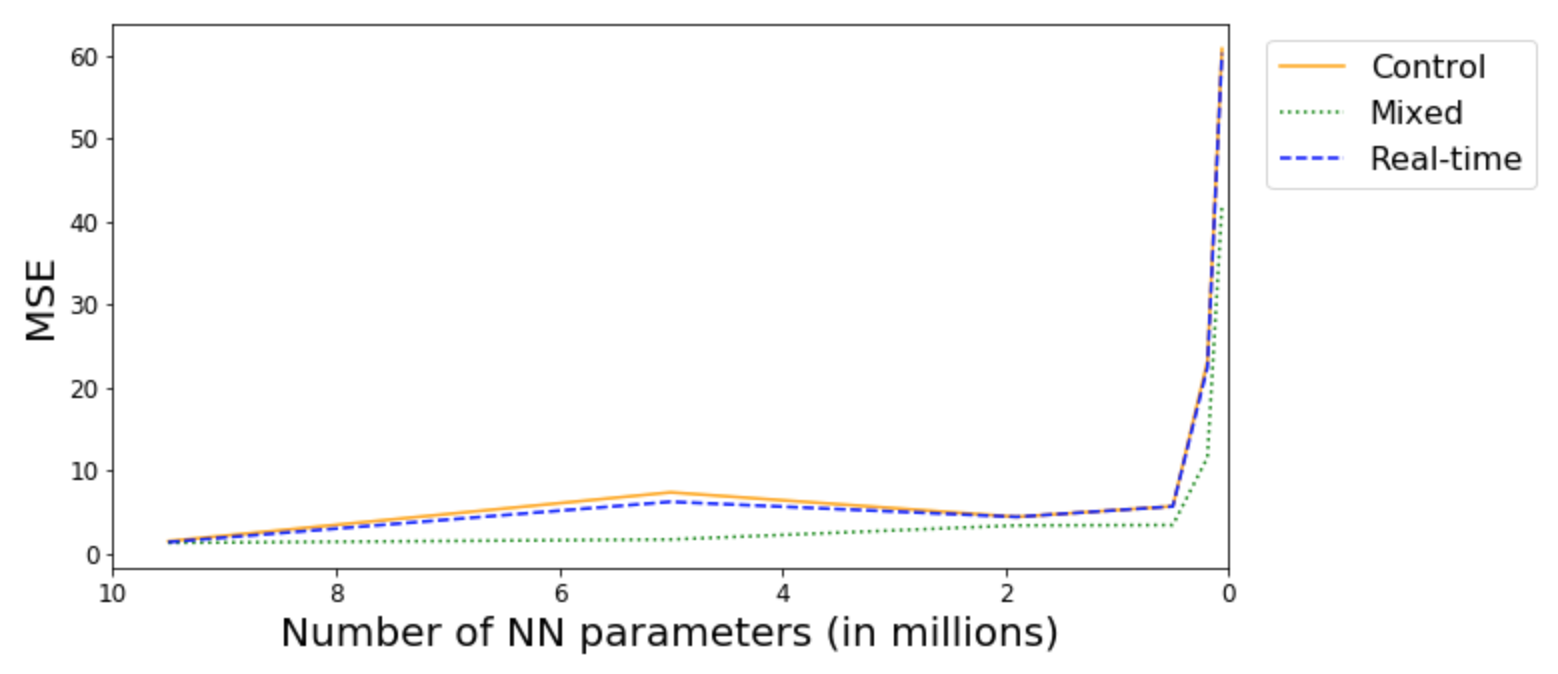}
    \caption{Train MSE} \label{fig:ReducedNNA}
\end{subfigure}%

\begin{subfigure}{\textwidth}
    \includegraphics[width=1\linewidth]{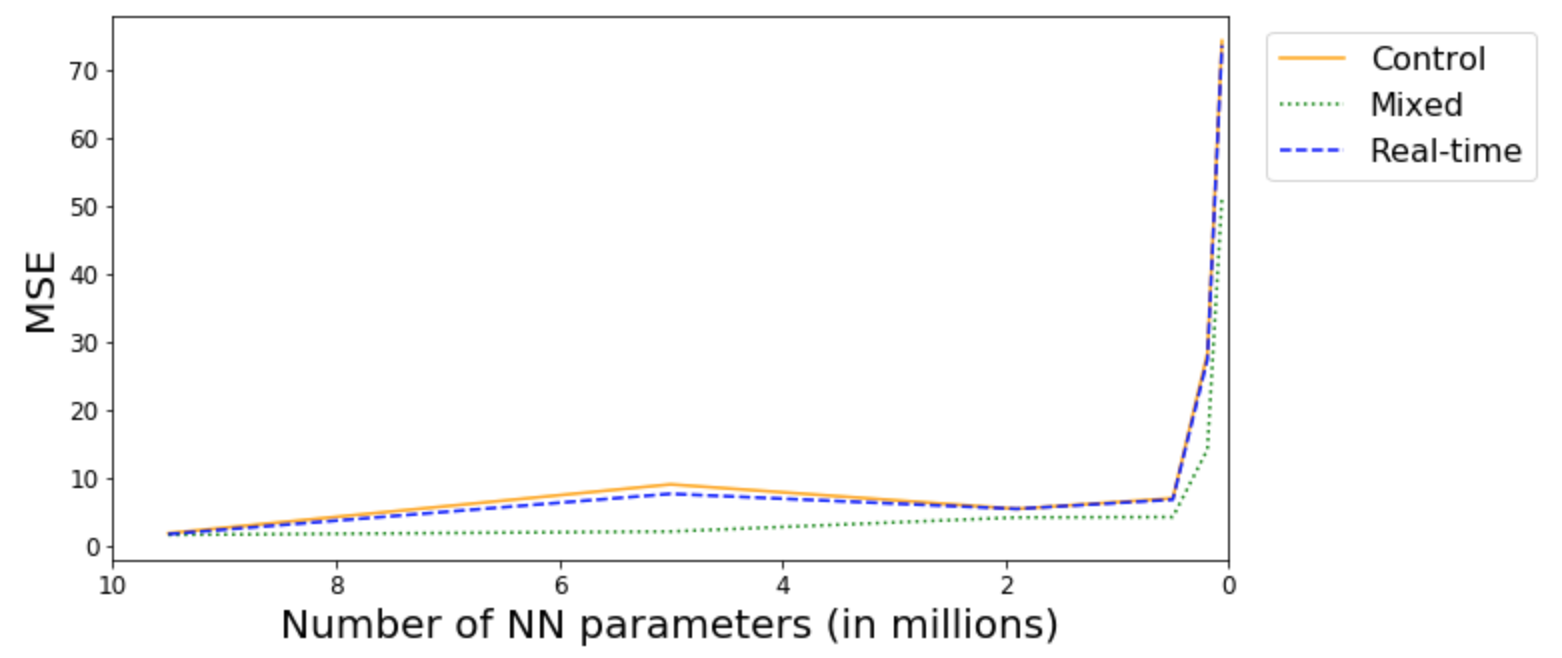}
    \caption{Validation MSE} \label{fig:ReducedNNB}
\end{subfigure}%

\begin{subfigure}{\textwidth}
    \includegraphics[width=1\linewidth]{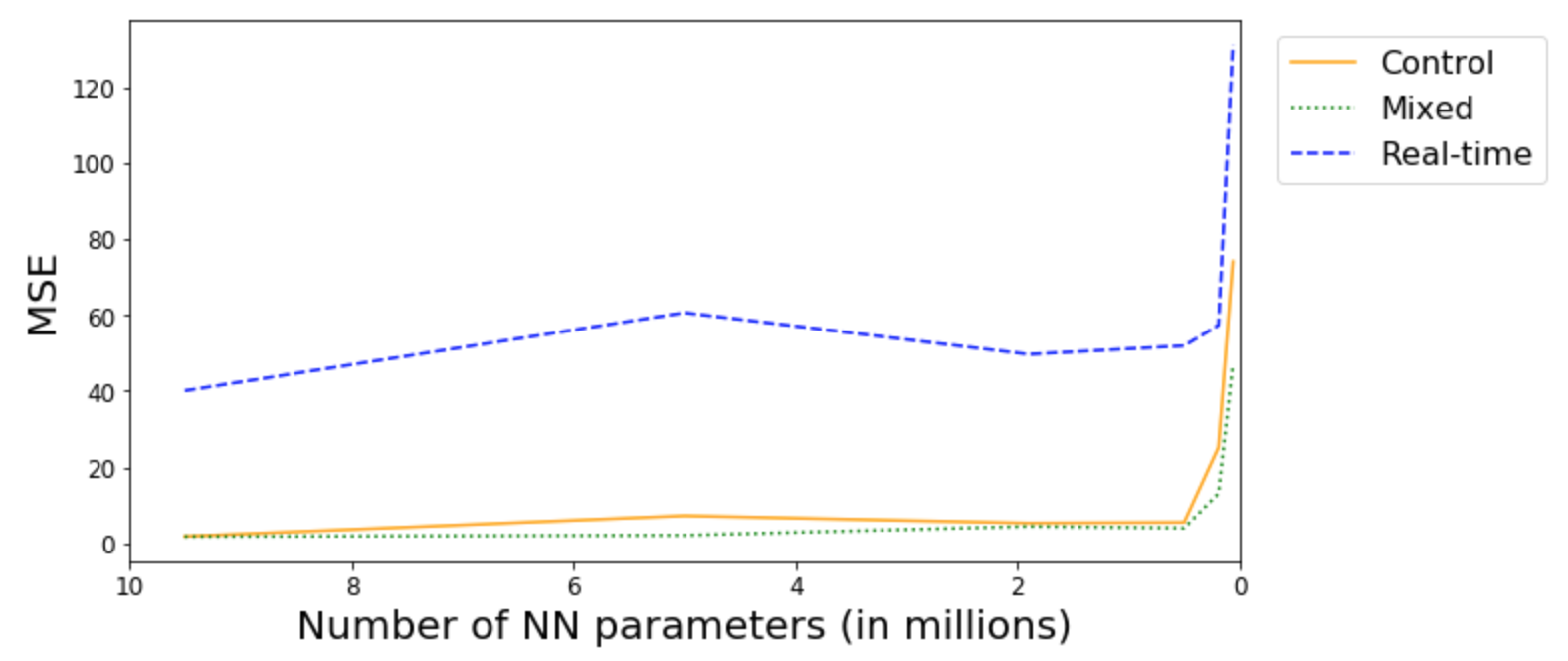}
    \caption{Test MSE} \label{fig:ReducedNNC}
\end{subfigure}%
\vspace{-3mm}
\caption{Average (a) Training, (b) Validation and (c) Testing error of Shallow NN on reduced subsets for various number of trainable parameters in three scenarios}
\label{fig:ReducedNN}
\end{figure}

\begin{figure}[!htbp]
\includegraphics[scale=0.25]{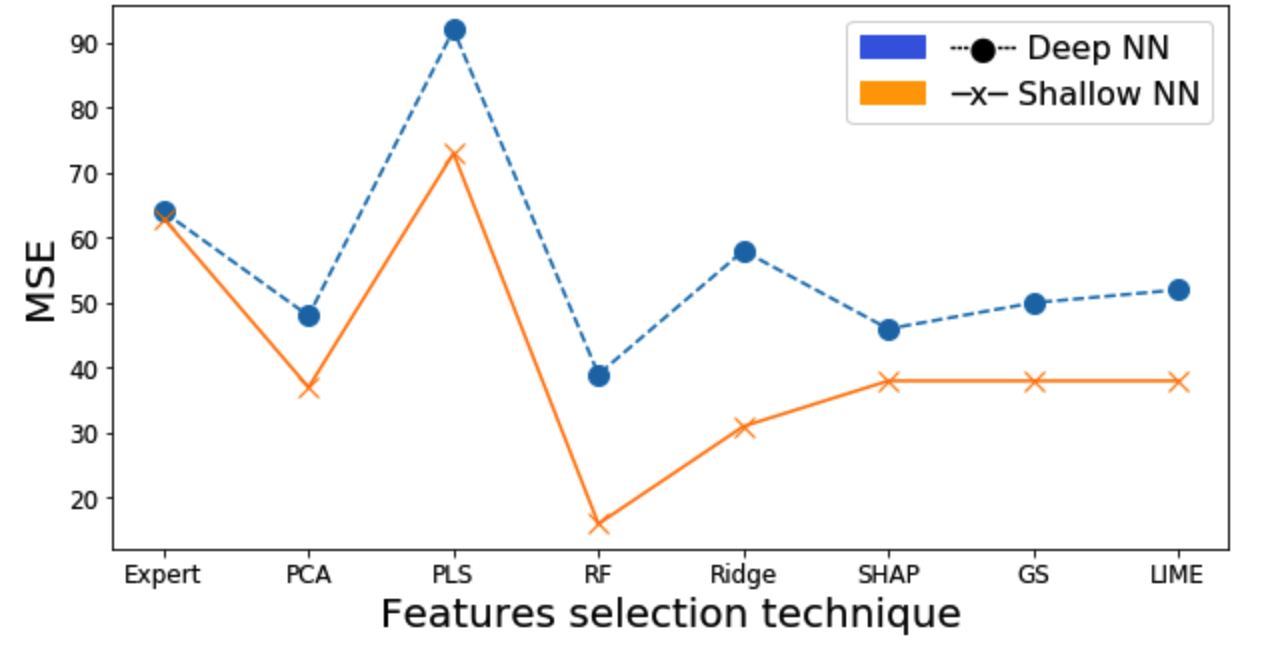}
    \caption{Shallow NN vs DNN testing error on reduced subsets in Real-time scenario.}
    \label{fig:ShallowVsDNN}
\end{figure}

Since we obtain two distinct architectures through BO, referred to as \emph{Shallow} for Full model and \emph{Deep} for Reduced model, we now cross-check their overall generalization ability. To this end, we employ heuristics on the constructed models, i.e., Shallow and Deep architectures for both Full and Reduced settings for the Real-time scenario (Table \ref{tab:NNgeneral}). As determined through optimization for the Reduced setting, the Shallow model performs better on unseen data than the Deep model for the Full setting.
In contrast, through an iterative reduction in the number of trainable parameters, based on the training (Figure \ref{fig:ReducedNNA}), validation (Figure \ref{fig:ReducedNNB}) and testing (Figure \ref{fig:ReducedNNC}) error, we observe that a less complex Shallow model (with only 0.5 million parameters) performed better in predicting unseen data observations compared to previously found ``optimal'' Reduced model with Deep architecture (Figure \ref{fig:ShallowVsDNN}). Such scaled model is also computationally more efficient (Figure \ref{fig:ComputeAandB}b). Hence, a more simple Shallow architecture proves to have better generalization ability for both Full and Reduced settings. Fine-tuned Reduced model with one hidden layer (472-954-1 nodes) is selected as the new optimal architecture.

\begin{table}[!htbp]
  \centering
    \begin{scriptsize}

    \begin{tabular}{|l|c|c|}
    \toprule
          & \textbf{Full Model} & \textbf{Reduced Model} \\
    \midrule
    \textbf{Shallow NN} & 32.3  & 42.5 \\
    \midrule
    \textbf{Depp NN} & 34.2  & 55.8 \\
    \bottomrule
    \end{tabular}
          \end{scriptsize}
    \caption{NN architecture generalization ability on Real-time data, recorded as Test MSE using different architectures on Full and Reduced sets of data}

    \label{tab:NNgeneral}%
\end{table}%


\subsection{Feature Selection for Spectra Data}
The following steps and tuning are performed to employ the considered feature selection and explainable AI techniques discussed in Section \S~\ref{sec:back} for spectroscopy analysis.

\begin{figure}[!htbp]
\includegraphics[scale=0.5]{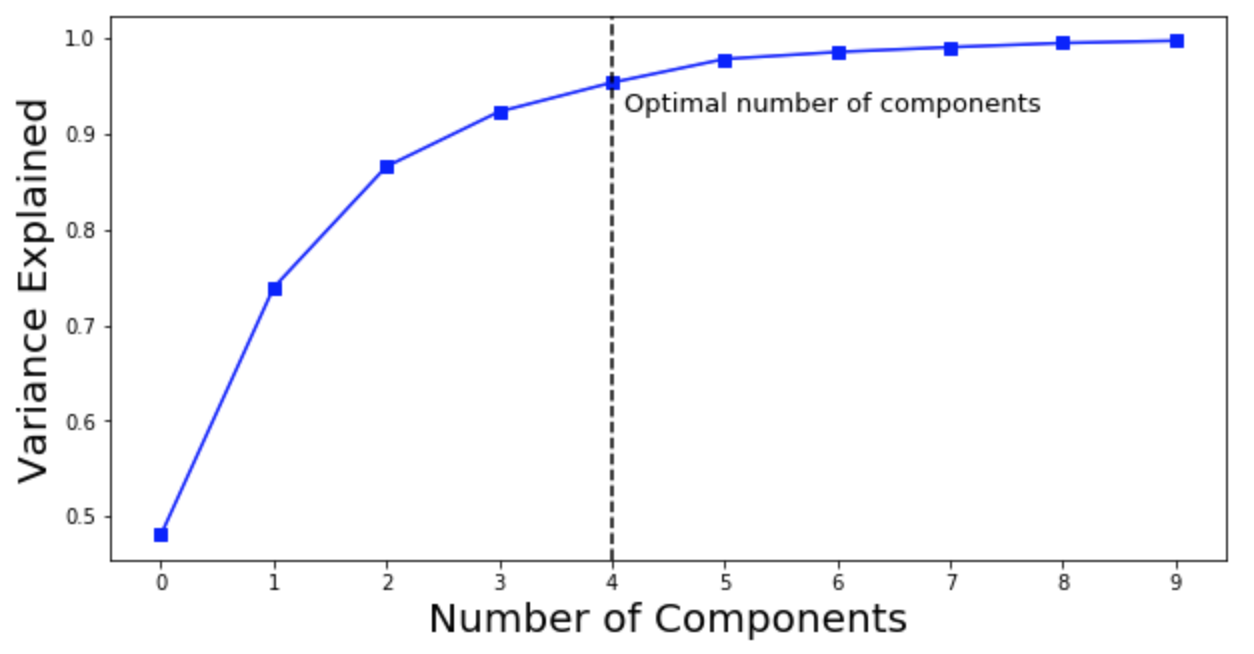}
    \caption{PCA cumulative sum of explained variance by number of components}
    \label{fig:PCA_comp}
\end{figure}

\begin{figure}[!htbp]
\includegraphics[scale=0.5]{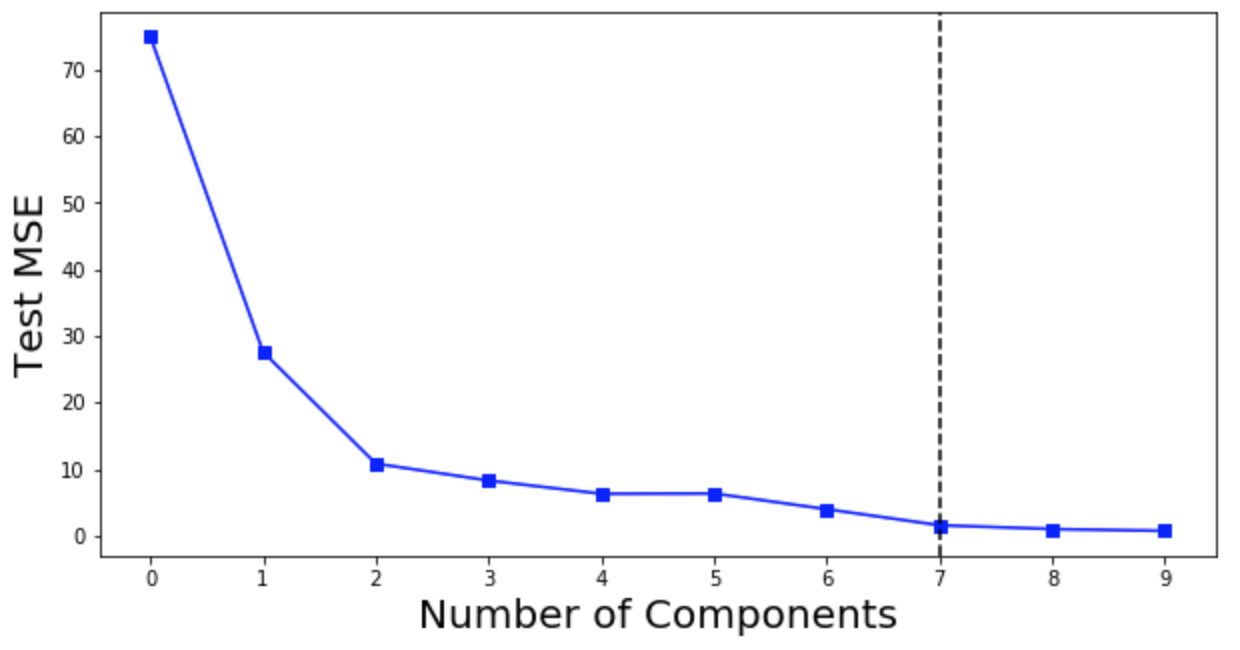}
    \caption{PLS Regression cross-decomposition and prediction results by number of components}
    \label{fig:PLS_comp}
\end{figure}

First, we select the optimal number of latent components for PCA and PLS by iteratively fitting spectroscopy data in scikit-learn corresponding decomposition libraries. 
For PCA, the number of components can be determined by plotting cumulative explained variance for $p$ components and selecting optimal number using point of maximum curvature (Figure \ref{fig:PCA_comp}). Since PLS considers the relationship between feature space and predicted output, we can determine an optimal number of components based on the uncertainty of test results for the different numbers of latent variables. Therefore, using PLS Regression, we fit the cross-decomposed data and record Mean Square Error (MSE) between predicted and true values of CN for a given number of components (Figure \ref{fig:PLS_comp}). Hence, the optimal number of latent components is selected based on the lowest associated MSE value. The optimal number of PCA components is determined to be $P=4$, and the optimal number of latent dimensions for PLS is determined to be $P=7$.


Consequently, we employ one additional step to determine the true contribution of each feature to the overall explained variance within the data. To accomplish that, we first calculate Explained Variance (EV), which is the ratio between the variance of that principal component and the total variance in data for PCA or predictions for PLS. We then take each component and multiply its EV with the loading vector to obtain individual feature importance.

\begin{equation}
    EV_{1} = 1-\frac{MSE_{1}}{\sum_{i}^{N}(y_{i}-\bar{y})^{2}})
\label{eq:EV1}
\end{equation}

\begin{equation}
    EV_{p+1} = EV_{p} - (1-\frac{MSE_{p+1}}{\sum_{i}^{N}(y_{i}-\bar{y})^{2}}),
\label{eq:EV2}
\end{equation}

The $sklearn$ toolbox allows us to directly access both the explained variance by component and loading vectors for PCA, while for PLS, we calculate EV of the first component $p=1$ using formula (\ref{eq:EV1}). For the remaining components $p=2,\dots, P$, the EV is calculated using Equation (\ref{eq:EV2}) to return the final cumulative explained variance by component normalized between 0 and 1. The features with the highest absolute weight values across all components are then aggregated and sorted. After filtering out features with zero weight, we achieve a true collection of selected features and their corresponding wavelengths that help explain most of the variance in the data.

To determine the optimal number of trees ($T$) to be used in the Random Forest Regression model and select the optimal $\alpha$ parameter for the Ridge Regression model, we use a similar approach as for PLS. We iteratively fit the data to the above-mentioned models and record MSE between predicted and true values of CN for given parameter $T$ in the range $[0.001, 1]$ and $\alpha$ in the range $[1,200]$. The optimal number of trees for the Random Forest Regression model is $T=91$. Similarly, for Ridge Regression we find optimal parameter value $\alpha=0.001$.

After the adjustments, 299 features are found to explain over 80\% of the total variance in training data using PCA, and 604 features are necessary using PLS, which is twice more than PCA selects. By extracting feature importance ($\mathbf{w}$) from the Random Forest model, we observe that 119 features account for 80\% of explained variance in data, significantly less than both PCA and PLS. As for Ridge Regression, 175 features with the highest corresponding linear weights ($\mathbf{w}$) explain most of the variance in the data.

We use the LimeTabular package\footnote{\url{https://github.com/marcotcr/lime}} to locally explain the behavior of optimized and trained NN and SVR models using Linear Regression as a simple model approximation. After perturbing the interpretable model input, features that contribute to individual fuel sample CN prediction are calculated, one observation at a time. In order to obtain a global explanation across the entire distribution of data, we record features with the highest associated coefficients (weights $w$) for each individual sample explanation using LR as a surrogate model. We then rank features selected most commonly across all sample instances and select them as the final subset of the most important features. 

The SHAP explainer\footnote{\url{https://github.com/slundberg/shap}} takes any combination of a predictive model and masker, which constrains the rules of the cooperative game used to explain the model and returns a callable subclass object that implements an estimation algorithm. We use SVR and NN prediction functions as input models to be explained and select a random subset of samples, generated by random seed function, to be used as a background (masking) set. By stratifying the background set on the model output results, instead of using the whole training data, we drastically reduce calculation time while maintaining a good representation of our sample distribution. For Global Surrogate\footnote{\url{https://github.com/interpretml/interpret-community}}, similar to LIME, we use an interpretable linear regression model to estimate general model behavior.

\begin{figure*}[!htbp]
\centering
\begin{subfigure}{1\textwidth}
    \includegraphics[width=1.2\linewidth,height=0.15\linewidth]{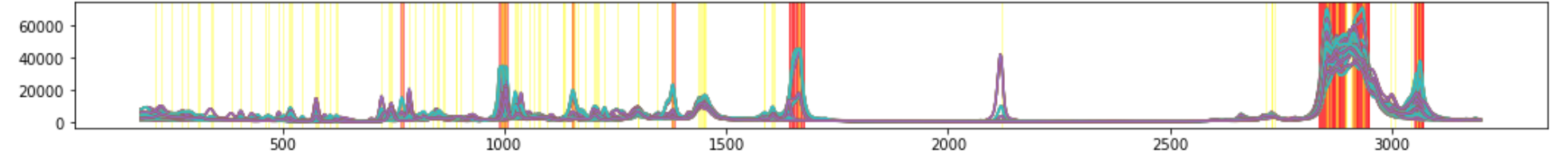}
    \caption{PCA}
\end{subfigure}%

\begin{subfigure}{\textwidth}
    \includegraphics[width=1.2\linewidth,height=0.15\linewidth]{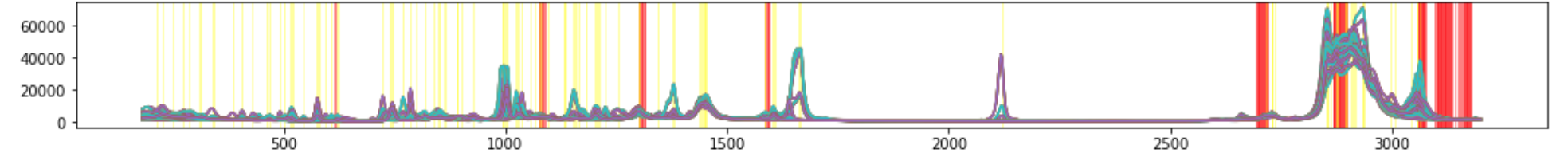}
    \caption{PLS}
\end{subfigure}%

\begin{subfigure}{\textwidth}
    \includegraphics[width=1.2\linewidth,height=0.15\linewidth]{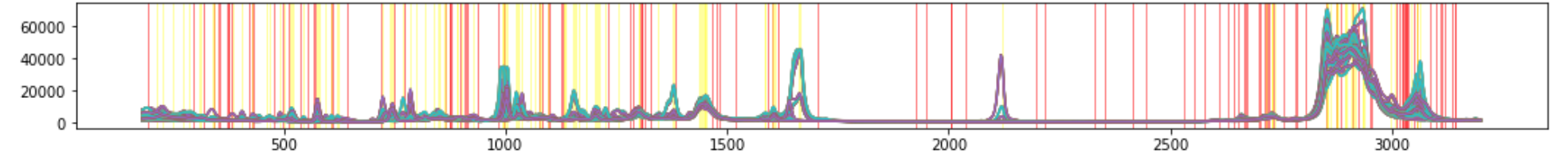}
    \caption{RF}
\end{subfigure}%

\begin{subfigure}{\textwidth}
    \includegraphics[width=1.2\linewidth,height=0.15\linewidth]{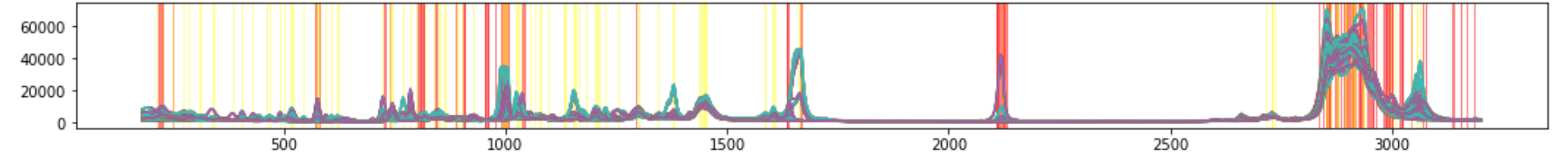}
    \caption{Ridge}
\end{subfigure}%

\begin{subfigure}{\textwidth}
    \includegraphics[width=1.2\linewidth,height=0.15\linewidth]{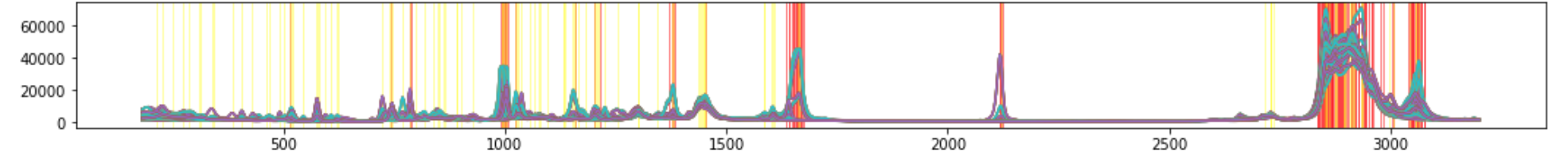}
    \caption{SHAP}
\end{subfigure}%

\begin{subfigure}{\textwidth}
    \includegraphics[width=1.2\linewidth,height=0.15\linewidth]{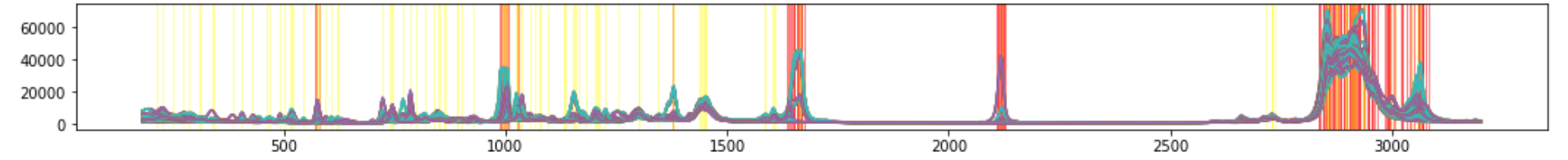}
    \caption{GS}
\end{subfigure}%

\begin{subfigure}{\textwidth}
    \includegraphics[width=1.2\linewidth,height=0.15\linewidth]{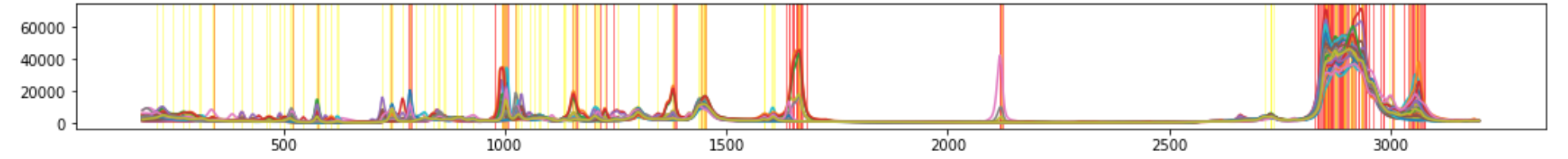}
    \caption{LIME}
\end{subfigure}%

\caption{Spectroscopy data for 145 sample Control scenario plotted. Regions of importance are marked as follows: Features selected by various techniques (a) PCA, (b) PLS, (c) Random Forest, (d) Ridge, (e) SHAPley, (f) Global Surrogate, (g) LIME are highlighted with \textit{red} and expert features are highlighted with \textit{yellow}.}
\label{Fig:FS}
\end{figure*}

\subsection{Evaluation Metrics}
\subsubsection{Correctness}

In this paper, the goal of feature selection is to select a subset of features in the spectra data that are considered most informative or relevant to the predicted CN values while adhering to known chemistry. The selected subset of features can be further compared to the theoretical locations of chemical functional groups within the spectroscopy data known from the domain expertise. Incorporating human knowledge in the learning loop allows us to evaluate the faithfulness of the applied techniques and ensure that a model can be trusted. 
To evaluate the performance of the feature selection techniques discussed in Section \S~\ref{sec:back}, we define a measure to calculate the alignment of the selected subset of features using different techniques with the selected features by the expert. We refer to this measure as \emph{correctness}. 

The selected wavenumbers are binned by unique centered intervals to represent their true location, which can be otherwise shifted by several wavenumbers due to instrument noise. The method Correctness is then calculated using simple Jaccard Similarity ($J$) \cite{jaccard1912distribution} between a given subset of binned features indicated by $S_{F}$, which is obtained from considered feature selection techniques, and expert-selected locations indicated by $S_{Ex}$. The method Correctness can also be measured by converting proportion $J$ to percentage to obtain values between 0 and 100\%, representing a percent match to all expert-selected locations. The Jaccard Similarity is calculated as follows:

\vspace{-2mm}
\begin{equation}
J(S_{F},S_{Ex}) = \dfrac{|S_{F}\cap{S_{Ex}}|}{|S_{F}\cup{S_{Ex}}|}=\dfrac{|S_{F}\cap{S_{Ex}}|}{|S_{F}|+|S_{Ex}|-|S_{F}\cap{S_{Ex}}|}
\end{equation}
\vspace{-3mm}

Although the performance of a predictive model after reducing the original feature space is important for at-scale implementation in real-time,  the correctness of the feature selection techniques is inevitable for transparency. In \S\ref{sec:results} we provide numerical results for the correctness of different considered techniques.

\subsubsection{Performance}
To aid the scalability and efficiency of the model deployment in practice, the model fitted on the reduced subset of features must be comparable in terms of performance to the model trained on full-spectrum and to a reduced model trained on an expert-selected subset of features. However, note that the reduced model is more efficient for the data collection process.
Since fuel CN is recorded and predicted as a real number, the performance of each model is evaluated using the Mean Squared Error (MSE) metric that calculates the deviation between the predicted and the true response values on the testing set. The MSE is formally represented as follows:
\vspace{-1mm}
\begin{equation}
MSE = \dfrac{1}{N}\sum\limits_{i=1}^{N}(y_i-\hat{y_i})^2 
\label{eq:MSE}
\end{equation}
\vspace{-3mm}

\section{Results}\label{sec:results}

Figure \ref{Fig:FS} illustrates the results of the considered feature selection techniques on our Control scenario dataset. The red bars indicate the selected features by the feature selection methods and the yellow bar inficates the features selected by the domain expert. As we can observe, the results indicate partial overlap of selected features with expert features around ``fingerprint'' region (500-1800 $cm^{-1}$) and ``Carbon-Hydrogen (CH) stretching'' (2800-3200 $cm^{-1}$) region, depending on the feature selection technique. Visual inspection suggests that 
similar subset of features is selected with model-agnostic methods.
Note that features selected by Random Forest are notably more uniformly spread across the entire spectrum than other approaches and have the least number of features from the CH stretching region. This behavior can be justified by the fact that Random Forest considers different subsets of features randomly selected for each of the individual models (Decision Trees) in its bag of models. Such considerations allows different attributes to be shown up in the tree structures and be considered for model construction. As a result, we can observe a wider range of locations selected as important in the spectra data. PLS and PCA results are more clustered in different regions. Both of these techniques are based on components that are linear combinations of original features. Hence, if attributes in one specific region are all important, they are assigned large loading values in multiple components and ranked higher in the final selected attributes.

\begin{figure}[!htbp]
\includegraphics[scale=0.5]{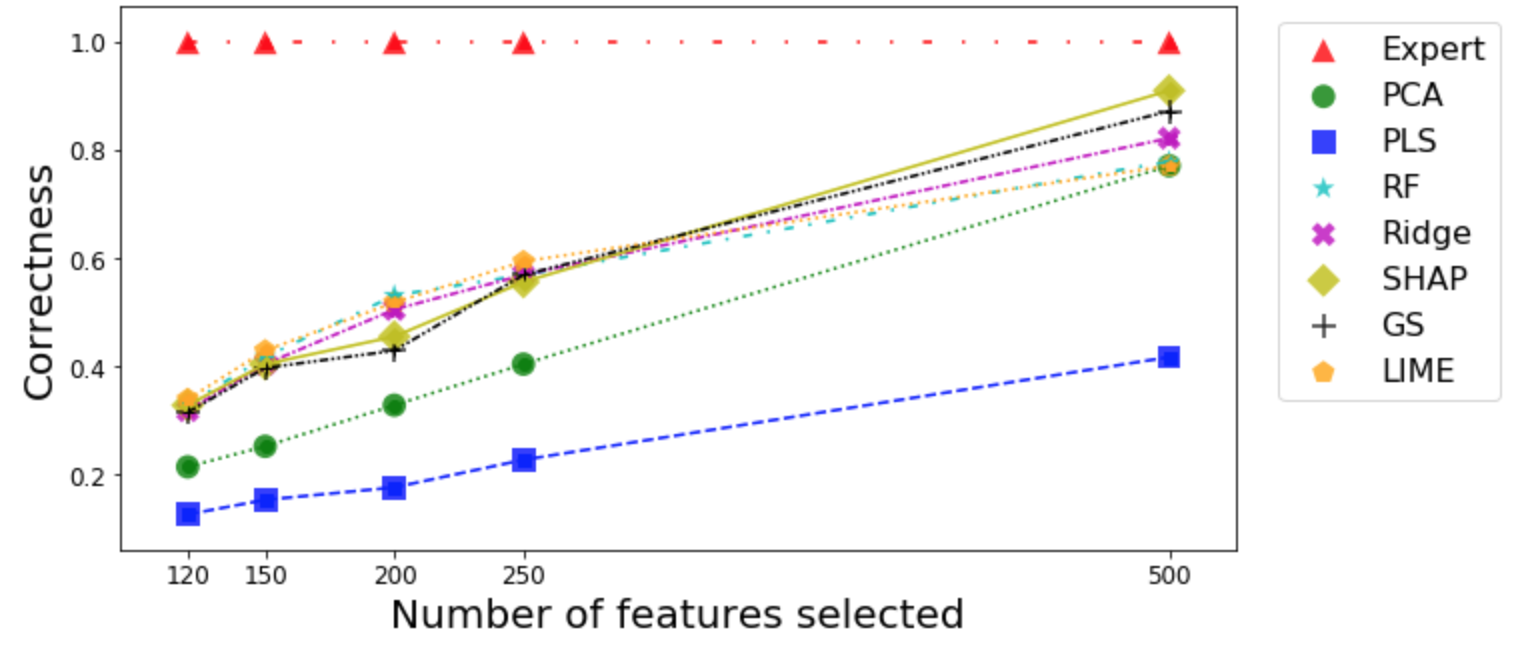}
\caption{Method Correctness, displayed as proportion of wavenumbers matched with the domain-defined features.}
\label{fig:Correctness}
\end{figure}

Figure \ref{fig:Correctness} represents the correctness evaluation of the considered techniques. The plot shows similarity proportion between subset of 120 Expert features $S_{Ex}$ and 120-500 features selected using various techniques $S_{F}$. We immediately observe that PCA and PLS are outperformed by other techniques in correlating to expert features even when we increase the upper bound on the number of selected features, averaging 17\%-60\% from 120 to 500 features. The results indicate that LIME, SHAP, Ridge, GS, and Random Forest feature selections performed competitively in model explainability resulting on average in 33\%-85\% of overlap with the selected features by the expert going from 120 to 500 features. 
Further, we notice that selecting 120 features from each method only partially overlaps with features selected by the expert. Recall that the exact locations of the majority of 120 expert-selected wavenumbers are rough estimations by the domain practitioners. Further, only a small subset of these features contribute to the variance in CN prediction. It is worth mentioning that the features selected by the considered feature selection methodologies are mainly returned, optimizing prediction performance. Hence, there are other locations than $S_{Ex}$ selected by such techniques. Since the expert selected features are based on rough peak location approximations, there might be actual locations that are missed. ML-based feature selection techniques might actually be able to discover the true locations of the functional groups' peaks. However, there is no solid approach to confirm this statement.

Table \ref{tab:finalResults} shows the comprehensive numerical prediction results of the considered Support Vector, and Neural Network regression models, as well as computational time. The NN model tuned through the Bayesian Optimization approach (BO-Tuned NN) results in a low testing MSE for Control and Mixed scenarios for both Full and Reduced sets of features. In a Real-time scenario, however, the minimum MSE is found to be $32.3$ and $39.5$ for Full and Reduced sets, accordingly. In contrast, after fine-tuning the NN, we can observe an increase in testing MSE for both Control and Mixed scenarios, while the error for Real-time scenarios becomes significantly smaller, achieving the new minimum MSE of $16$ in the Reduced setting using Random Forest. Using the BO-Tuned SVR model, we obtain more competitive testing results across all three scenarios, achieving the lowest MSE of $33.7$ for the Full set and even lower $32.9$ for the Reduced set of features. Upon further adjusting the SVR margin parameter, we notice a slight increase in testing error for Control and Mixed scenarios while achieving all-time low MSE for Real-time scenarios of $14.8$ for Full and $4.6$ for the Reduced set of attributes selected by Random Forest. While not included in the table, using Linear Regression, we achieve the lowest MSE of $88.8$ for the Full model and $233$ for the Reduced model, using the Random Forest subset of attributes.

\begin{table*}[!htbp]
  \centering
  \tiny
    \begin{tabular}{|rr|ccc|ccc|cccr|}
    \toprule
    \multicolumn{1}{|c|}{\multirow{2}[2]{*}{\textbf{Number of samples}}} & \multicolumn{3}{c|}{\multirow{2}[2]{*}{\textbf{145}}} & \multicolumn{3}{c|}{\multirow{2}[2]{*}{\textbf{245}}} & \multicolumn{3}{c|}{\multirow{2}[2]{*}{\textbf{165}}} &       &  \\
    \multicolumn{1}{|c|}{} & \multicolumn{3}{c|}{} & \multicolumn{3}{c|}{} & \multicolumn{3}{c|}{} &       &  \\
    \midrule
    \multicolumn{1}{|c|}{\multirow{2}[4]{*}{\textbf{FS method}}} & \multicolumn{3}{c|}{\textbf{Control}} & \multicolumn{3}{c|}{\textbf{Mixed}} & \multicolumn{3}{c|}{\textbf{Real-time}} & \multicolumn{1}{c|}{} & \multicolumn{1}{c|}{\multirow{11}[22]{*}{\begin{sideways}\textbf{BO-Tuned NN}\end{sideways}}} \\
\cmidrule{2-11}    \multicolumn{1}{|c|}{} & \multicolumn{1}{c|}{Train ($\mu$,$\sigma$)} & \multicolumn{1}{c|}{\textbf{Test ($\mu$,$\sigma$)}} & \multicolumn{1}{c|}{Time (s)} & Train ($\mu$,$\sigma$) & \multicolumn{1}{c|}{\textbf{Test ($\mu$,$\sigma$)}} & \multicolumn{1}{c|}{Time (s)} & Train ($\mu$,$\sigma$) & \multicolumn{1}{c|}{\textbf{Test ($\mu$,$\sigma$)}} & \multicolumn{1}{c|}{Time (s)} & \multicolumn{1}{c|}{\# parameters} &  \\
\cmidrule{1-11}    \multicolumn{1}{|l|}{Full Model} & \multicolumn{1}{c|}{(0.006, 0.004)} & \multicolumn{1}{c|}{(0.04, 0.04)} & \multicolumn{1}{c|}{2338} & (0.02, 0.05) & \multicolumn{1}{c|}{(0.07, 0.13)} & \multicolumn{1}{c|}{2404} & (0.01, 0.01) & \multicolumn{1}{c|}{{(32.3, 6.7)}} & \multicolumn{1}{c|}{2533} & \multicolumn{1}{c|}{11.8 mil} &  \\
\cmidrule{1-11}    \multicolumn{1}{|l|}{Expert } & \multicolumn{1}{c|}{(0.02, 0.006)} & \multicolumn{1}{c|}{(0.04, 0.02)} & \multicolumn{1}{c|}{3506} & (0.02, 0.03) & \multicolumn{1}{c|}{(0.03, 0.15)} & \multicolumn{1}{c|}{4170} & (0.03, 0.02) & \multicolumn{1}{c|}{(64, 19)} & \multicolumn{1}{c|}{2922} & \multicolumn{1}{c|}{13.2 mil} &  \\
\cmidrule{1-11}    \multicolumn{1}{|l|}{PCA} & \multicolumn{1}{c|}{(0.05, 0.02)} & \multicolumn{1}{c|}{(0.07, 0.03)} & \multicolumn{1}{c|}{3607} & (0.08, 0.02 & \multicolumn{1}{c|}{(0.17, 0.06)} & \multicolumn{1}{c|}{4230} & (0.04, 0.02) & \multicolumn{1}{c|}{(48, 20)} & \multicolumn{1}{c|}{2439} & \multicolumn{1}{c|}{13.2 mil} &  \\
\cmidrule{1-11}    \multicolumn{1}{|l|}{PLS} & \multicolumn{1}{c|}{(0.29, 0.18)} & \multicolumn{1}{c|}{(0.5, 0.58)} & \multicolumn{1}{c|}{3896} & (2.07, 0.85) & \multicolumn{1}{c|}{(1.6, 1)} & \multicolumn{1}{c|}{4832} & (1.66, 0.7) & \multicolumn{1}{c|}{(92, 17.2)} & \multicolumn{1}{c|}{2835} & \multicolumn{1}{c|}{13.2 mil} &  \\
\cmidrule{1-11}    \multicolumn{1}{|l|}{RF} & \multicolumn{1}{c|}{(0.05, 0.01)} & \multicolumn{1}{c|}{(0.08, 0.03)} & \multicolumn{1}{c|}{3421} & (0.04, 0.02) & \multicolumn{1}{c|}{(0.06, 0.02)} & \multicolumn{1}{c|}{4260} & (0.09, 0.08) & \multicolumn{1}{c|}{{(39.5, 12.8)}} & \multicolumn{1}{c|}{2421} & \multicolumn{1}{c|}{13.2 mil} &  \\
\cmidrule{1-11}    \multicolumn{1}{|l|}{Ridge} & \multicolumn{1}{c|}{(0.05, 0.03)} & \multicolumn{1}{c|}{(0.07, 0.03)} & \multicolumn{1}{c|}{3327} & (0.05, 0.01) & \multicolumn{1}{c|}{(0.07, 0.03)} & \multicolumn{1}{c|}{4122} & (0.06, 0.07) & \multicolumn{1}{c|}{(58.5, 20.6)} & \multicolumn{1}{c|}{2855} & \multicolumn{1}{c|}{13.2 mil} &  \\
\cmidrule{1-11}    \multicolumn{1}{|l|}{SHAP } & \multicolumn{1}{c|}{(0.04, 0.02)} & \multicolumn{1}{c|}{(0.05, 0.02)} & \multicolumn{1}{c|}{3266} & (0.02, 0.01) & \multicolumn{1}{c|}{{(0.02, 0.01)}} & \multicolumn{1}{c|}{4620} & (0.04, 0.025) & \multicolumn{1}{c|}{(46, 32)} & \multicolumn{1}{c|}{2806} & \multicolumn{1}{c|}{13.2 mil} &  \\
\cmidrule{1-11}    \multicolumn{1}{|l|}{GS } & \multicolumn{1}{c|}{(0.04, 0.01)} & \multicolumn{1}{c|}{(0.06, 0.03)} & \multicolumn{1}{c|}{3285} & (0.04, 0.02) & \multicolumn{1}{c|}{(0.06, 0.02)} & \multicolumn{1}{c|}{4410} & (0.03, 0.027) & \multicolumn{1}{c|}{(50, 12)} & \multicolumn{1}{c|}{2460} & \multicolumn{1}{c|}{13.2 mil} &  \\
\cmidrule{1-11}    \multicolumn{1}{|l|}{LIME } & \multicolumn{1}{c|}{(0.03, 0.01)} & \multicolumn{1}{c|}{(0.05, 0.03)} & \multicolumn{1}{c|}{3217} & (0.05, 0.02) & \multicolumn{1}{c|}{(0.09, 0.03)} & \multicolumn{1}{c|}{4350} & (0.06, 0.035) & \multicolumn{1}{c|}{(52, 14)} & \multicolumn{1}{c|}{2444} & \multicolumn{1}{c|}{13.2 mil} &  \\
    \midrule
    \textbf{Avg. runtime} & \multicolumn{1}{r}{} &       & \textbf{111} &       &       & \textbf{139} &       &       & \textbf{88} &       &  \\
    \midrule
    \multicolumn{1}{|l|}{Full Model} & \multicolumn{1}{c|}{(0.006, 0.004)} & \multicolumn{1}{c|}{(0.04, 0.04)} & \multicolumn{1}{c|}{2404} & (0.024, 0.045) & \multicolumn{1}{c|}{{(0.07, 0.127)}} & \multicolumn{1}{c|}{2617} & (0.01, 0.01) & \multicolumn{1}{c|}{(32.3, 6.7)} & \multicolumn{1}{c|}{2613} & \multicolumn{1}{c|}{11.8 mil} & \multicolumn{1}{c|}{\multirow{9}[18]{*}{\begin{sideways}\textbf{Fine-Tuned NN}\end{sideways}}} \\
\cmidrule{1-11}    \multicolumn{1}{|l|}{Expert Features} & \multicolumn{1}{c|}{(0.05, 0.03)} & \multicolumn{1}{c|}{(0.25, 0.2)} & \multicolumn{1}{c|}{344} & (0.19, 0.35) & \multicolumn{1}{c|}{(0.26, 0.5)} & \multicolumn{1}{c|}{375} & (5, 1) & \multicolumn{1}{c|}{(63.7, 3)} & \multicolumn{1}{c|}{351} & \multicolumn{1}{c|}{0.5 mil} &  \\
\cmidrule{1-11}    \multicolumn{1}{|l|}{PCA} & \multicolumn{1}{c|}{(1.3, 0.3)} & \multicolumn{1}{c|}{(4.9, 2.5)} & \multicolumn{1}{c|}{380} & (0.5, 0.83) & \multicolumn{1}{c|}{(0.67, 1.1)} & \multicolumn{1}{c|}{391} & (0.21, 0.5) & \multicolumn{1}{c|}{(42.6, 11)} & \multicolumn{1}{c|}{381} & \multicolumn{1}{c|}{0.5 mil} &  \\
\cmidrule{1-11}    \multicolumn{1}{|l|}{PLS} & \multicolumn{1}{c|}{(22, 15)} & \multicolumn{1}{c|}{(19, 20)} & \multicolumn{1}{c|}{391} & (12.3, 1.9) & \multicolumn{1}{c|}{(13.9, 4.6)} & \multicolumn{1}{c|}{462} & (11.7, 0.68) & \multicolumn{1}{c|}{(73.8, 7.2)} & \multicolumn{1}{c|}{366} & \multicolumn{1}{c|}{0.5 mil} &  \\
\cmidrule{1-11}    \multicolumn{1}{|l|}{RF} & \multicolumn{1}{c|}{(0.86, 1.15)} & \multicolumn{1}{c|}{(1.62, 1.9)} & \multicolumn{1}{c|}{377} & (4.6, 0.8) & \multicolumn{1}{c|}{(5.7, 1.9)} & \multicolumn{1}{c|}{358} & (0.73, 1.32) & \multicolumn{1}{c|}{{(16, 1)}} & \multicolumn{1}{c|}{456} & \multicolumn{1}{c|}{0.5 mil} &  \\
\cmidrule{1-11}    \multicolumn{1}{|l|}{Ridge} & \multicolumn{1}{c|}{(0.73, 0.96)} & \multicolumn{1}{c|}{(2.7, 3.8)} & \multicolumn{1}{c|}{361} & (0.6, 0.96) & \multicolumn{1}{c|}{(1.03, 1.95)} & \multicolumn{1}{c|}{377} & (3, 0.34) & \multicolumn{1}{c|}{(31.7, 2)} & \multicolumn{1}{c|}{396} & \multicolumn{1}{c|}{0.5 mil} &  \\
\cmidrule{1-11}    \multicolumn{1}{|l|}{SHAP } & \multicolumn{1}{c|}{(0.001, 0.0015)} & \multicolumn{1}{c|}{(0.06, 0.13)} & \multicolumn{1}{c|}{384} & (0.67, 0.87) & \multicolumn{1}{c|}{(0.82, 0.996)} & \multicolumn{1}{c|}{470} & (3, 0.4) & \multicolumn{1}{c|}{(38.6, 0.9)} & \multicolumn{1}{c|}{357} & \multicolumn{1}{c|}{0.5 mil} &  \\
\cmidrule{1-11}    \multicolumn{1}{|l|}{GS } & \multicolumn{1}{c|}{(0.001, 0.008)} & \multicolumn{1}{c|}{(0.02, 0.029)} & \multicolumn{1}{c|}{362} & (0.83, 1.19) & \multicolumn{1}{c|}{(0.83, 1.07)} & \multicolumn{1}{c|}{383} & (0.35, 0.63) & \multicolumn{1}{c|}{(38, 2.1)} & \multicolumn{1}{c|}{369} & \multicolumn{1}{c|}{0.5 mil} &  \\
\cmidrule{1-11}    \multicolumn{1}{|l|}{LIME } & \multicolumn{1}{c|}{(0.048, 0.07)} & \multicolumn{1}{c|}{(0.225, 0.18)} & \multicolumn{1}{c|}{366} & (0.42, 0.72) & \multicolumn{1}{c|}{(0.5, 0.67)} & \multicolumn{1}{c|}{376} & (0.13, 0.43) & \multicolumn{1}{c|}{(38, 2.3)} & \multicolumn{1}{c|}{333} & \multicolumn{1}{c|}{0.5 mil} &  \\
    \midrule
    \textbf{Avg. runtime} &       &       & \textbf{19.89} &       &       & \textbf{21.51} &       &       & \textbf{20.82} &       &  \\
    \midrule
    \multicolumn{1}{|l|}{Full Model} & \multicolumn{1}{c|}{(0.006, 0.002)} & \multicolumn{1}{c|}{(0.02, 0.008)} & \multicolumn{1}{c|}{0.58} & (0.007, 0.0002) & \multicolumn{1}{c|}{{(0.05, 0.019)}} & \multicolumn{1}{c|}{11.4} & (0.006, 0.002) & \multicolumn{1}{c|}{(33.67, 5.44)} & \multicolumn{1}{c|}{0.63} & \multicolumn{1}{c|}{} & \multicolumn{1}{c|}{\multirow{9}[18]{*}{\begin{sideways}\textbf{BO-Tuned SVR}\end{sideways}}} \\
\cmidrule{1-11}    \multicolumn{1}{|l|}{Expert Features} & \multicolumn{1}{c|}{(0.006, 0.002)} & \multicolumn{1}{c|}{(0.02, 0.009)} & \multicolumn{1}{c|}{1.2} & (0.01, 0.007) & \multicolumn{1}{c|}{(0.05, 0.02)} & \multicolumn{1}{c|}{35.97} & (0.006, 0.002) & \multicolumn{1}{c|}{(33.95, 6.4)} & \multicolumn{1}{c|}{1.2} & \multicolumn{1}{c|}{} &  \\
\cmidrule{1-11}    \multicolumn{1}{|l|}{PCA} & \multicolumn{1}{c|}{(0.007, 0.002)} & \multicolumn{1}{c|}{(0.04, 0.015)} & \multicolumn{1}{c|}{2.75} & (0.026, 0.04) & \multicolumn{1}{c|}{(0.1, 0.08)} & \multicolumn{1}{c|}{42.25} & (0.007, 0.002) & \multicolumn{1}{c|}{(34, 5.5)} & \multicolumn{1}{c|}{2.65} & \multicolumn{1}{c|}{} &  \\
\cmidrule{1-11}    \multicolumn{1}{|l|}{PLS} & \multicolumn{1}{c|}{(0.016, 0.007)} & \multicolumn{1}{c|}{(0.32, 0.115)} & \multicolumn{1}{c|}{30.8} & (1.3, 1.5) & \multicolumn{1}{c|}{(2.46, 2.3)} & \multicolumn{1}{c|}{61.12} & (0.016, 0.007) & \multicolumn{1}{c|}{(42.5, 9.2)} & \multicolumn{1}{c|}{30.27} & \multicolumn{1}{c|}{} &  \\
\cmidrule{1-11}    \multicolumn{1}{|l|}{RF} & \multicolumn{1}{c|}{(0.008, 0.002)} & \multicolumn{1}{c|}{(0.06, 0.03)} & \multicolumn{1}{c|}{5.2} & (0.8, 0.74) & \multicolumn{1}{c|}{(0.36, 0.22)} & \multicolumn{1}{c|}{56.78} & (0.008, 0.002) & \multicolumn{1}{c|}{(41.2, 2.1)} & \multicolumn{1}{c|}{5.35} & \multicolumn{1}{c|}{} &  \\
\cmidrule{1-11}    \multicolumn{1}{|l|}{Ridge} & \multicolumn{1}{c|}{(0.006, 0.003)} & \multicolumn{1}{c|}{(0.03, 0.013)} & \multicolumn{1}{c|}{1.3} & (0.07, 0.07) & \multicolumn{1}{c|}{(0.13, 0.08)} & \multicolumn{1}{c|}{41.97} & (0.007, 0.003) & \multicolumn{1}{c|}{(37.9, 9.1)} & \multicolumn{1}{c|}{1.33} & \multicolumn{1}{c|}{} &  \\
\cmidrule{1-11}    \multicolumn{1}{|l|}{SHAP } & \multicolumn{1}{c|}{(0.006, 0.002)} & \multicolumn{1}{c|}{(0.03, 0.01)} & \multicolumn{1}{c|}{1.8} & (0.11, 0.3) & \multicolumn{1}{c|}{(0.46, 1.83)} & \multicolumn{1}{c|}{39.84} & (0.006, 0.002) & \multicolumn{1}{c|}{{(32.85, 4.7)}} & \multicolumn{1}{c|}{1.83} & \multicolumn{1}{c|}{} &  \\
\cmidrule{1-11}    \multicolumn{1}{|l|}{GS } & \multicolumn{1}{c|}{(0.006, 0.002)} & \multicolumn{1}{c|}{(0.03, 0.013)} & \multicolumn{1}{c|}{1.7} & (0.14, 0.58) & \multicolumn{1}{c|}{(0.44, 1.67)} & \multicolumn{1}{c|}{38.93} & (0.007, 0.003) & \multicolumn{1}{c|}{(36, 6.3)} & \multicolumn{1}{c|}{1.63} & \multicolumn{1}{c|}{} &  \\
\cmidrule{1-11}    \multicolumn{1}{|l|}{LIME } & \multicolumn{1}{c|}{(0.006, 0.002)} & \multicolumn{1}{c|}{(0.02, 0.008)} & \multicolumn{1}{c|}{1} & (0.027, 0.06) & \multicolumn{1}{c|}{{(0.08, 0.05)}} & \multicolumn{1}{c|}{37.63} & (0.006, 0.002) & \multicolumn{1}{c|}{(36, 3.74)} & \multicolumn{1}{c|}{1.06} & \multicolumn{1}{c|}{} &  \\
    \midrule
    \textbf{Avg. runtime} &       &       & \textbf{0.17} &       &       & \textbf{1.4} &       &       & \textbf{0.2} &       &  \\
    \midrule
    \multicolumn{1}{|l|}{Full Model} & \multicolumn{1}{c|}{(0.07, 0.001)} & \multicolumn{1}{c|}{(0.08, 0.02)} & \multicolumn{1}{c|}{0.41} & (0.007, 0.0002) & \multicolumn{1}{c|}{{(0.11, 0.03)}} & \multicolumn{1}{c|}{5.18} & (0.006, 0.002) & \multicolumn{1}{c|}{(14.8, 1.1)} & \multicolumn{1}{c|}{0.45} & \multicolumn{1}{c|}{} & \multicolumn{1}{c|}{\multirow{9}[18]{*}{\begin{sideways}\textbf{Fine-Tuned SVR}\end{sideways}}} \\
\cmidrule{1-11}    \multicolumn{1}{|l|}{Expert Features} & \multicolumn{1}{c|}{(0.06, 0.001)} & \multicolumn{1}{c|}{(0.08, 0.02)} & \multicolumn{1}{c|}{0.36} & (0.06, 0.001) & \multicolumn{1}{c|}{(0.11, 0.04)} & \multicolumn{1}{c|}{17.2} & (0.006, 0.002) & \multicolumn{1}{c|}{(14, 2.1)} & \multicolumn{1}{c|}{0.36} & \multicolumn{1}{c|}{} &  \\
\cmidrule{1-11}    \multicolumn{1}{|l|}{PCA} & \multicolumn{1}{c|}{(0.07, 0.003)} & \multicolumn{1}{c|}{(0.11, 0.03)} & \multicolumn{1}{c|}{0.97} & (0.06, 0.002) & \multicolumn{1}{c|}{(0.1, 0.08)} & \multicolumn{1}{c|}{24.1} & (0.007, 0.002) & \multicolumn{1}{c|}{(15.2, 1.2)} & \multicolumn{1}{c|}{0.98} & \multicolumn{1}{c|}{} &  \\
\cmidrule{1-11}    \multicolumn{1}{|l|}{PLS} & \multicolumn{1}{c|}{(0.07, 0.003)} & \multicolumn{1}{c|}{(0.13, 0.04)} & \multicolumn{1}{c|}{5.1} & (1, 0.79) & \multicolumn{1}{c|}{(1.5, 0.9)} & \multicolumn{1}{c|}{39.2} & (0.016, 0.007) & \multicolumn{1}{c|}{(16.2, 1.4)} & \multicolumn{1}{c|}{5} & \multicolumn{1}{c|}{} &  \\
\cmidrule{1-11}    \multicolumn{1}{|l|}{RF} & \multicolumn{1}{c|}{(0.06, 0.002)} & \multicolumn{1}{c|}{(0.08, 0.03)} & \multicolumn{1}{c|}{0.49} & (0.06, 0.001) & \multicolumn{1}{c|}{(0.09, 0.02)} & \multicolumn{1}{c|}{5.5} & (0.008, 0.002) & \multicolumn{1}{c|}{{(4.6, 0.2)}} & \multicolumn{1}{c|}{0.5} & \multicolumn{1}{c|}{} &  \\
\cmidrule{1-11}    \multicolumn{1}{|l|}{Ridge} & \multicolumn{1}{c|}{(0.06, 0.002)} & \multicolumn{1}{c|}{(0.08, 0.02)} & \multicolumn{1}{c|}{0.37} & (0.06, 0.001) & \multicolumn{1}{c|}{(0.1, 0.03)} & \multicolumn{1}{c|}{7.9} & (0.007, 0.003) & \multicolumn{1}{c|}{(11.8, 0.7)} & \multicolumn{1}{c|}{0.37} & \multicolumn{1}{c|}{} &  \\
\cmidrule{1-11}    \multicolumn{1}{|l|}{SHAP } & \multicolumn{1}{c|}{(0.07, 0.003)} & \multicolumn{1}{c|}{(0.1, 0.03)} & \multicolumn{1}{c|}{0.7} & (0.06, 0.002) & \multicolumn{1}{c|}{(0.14, 0.05)} & \multicolumn{1}{c|}{22.3} & (0.006, 0.002) & \multicolumn{1}{c|}{(12.5, 0.7)} & \multicolumn{1}{c|}{0.73} & \multicolumn{1}{c|}{} &  \\
\cmidrule{1-11}    \multicolumn{1}{|l|}{GS } & \multicolumn{1}{c|}{(0.06, 0.002)} & \multicolumn{1}{c|}{(0.1, 0.02)} & \multicolumn{1}{c|}{0.68} & (0.06, 0.002) & \multicolumn{1}{c|}{(0.12, 0.04)} & \multicolumn{1}{c|}{16.1} & (0.007, 0.003) & \multicolumn{1}{c|}{(11.8, 1.1)} & \multicolumn{1}{c|}{0.7} & \multicolumn{1}{c|}{} &  \\
\cmidrule{1-11}    \multicolumn{1}{|l|}{LIME } & \multicolumn{1}{c|}{(0.06, 0.002)} & \multicolumn{1}{c|}{(0.1, 0.02)} & \multicolumn{1}{c|}{0.56} & (0.06, 0.002) & \multicolumn{1}{c|}{{(0.12, 0.03)}} & \multicolumn{1}{c|}{14.7} & (0.006, 0.002) & \multicolumn{1}{c|}{(13.9, 0.9)} & \multicolumn{1}{c|}{0.55} & \multicolumn{1}{c|}{} &  \\
    \midrule
    \textbf{Avg. runtime} &       &       & \textbf{0.04} &       &       & \textbf{0.6} &       &       & \textbf{0.04} &       &  \\

    \bottomrule
    \end{tabular}%
\caption{Performance of NN, SVR and LR predictive models fitted with Full and Reduced sets of features measured using mean training and testing error ($\mu$ MSE) with corresponding mean deviation across 30 random splits of data ($\sigma$). Average computation time for training one model summarized below the performance results for each scenario.}
\label{tab:finalResults}%
\end{table*}%

\begin{figure}[!htbp]
\includegraphics[scale=0.25]{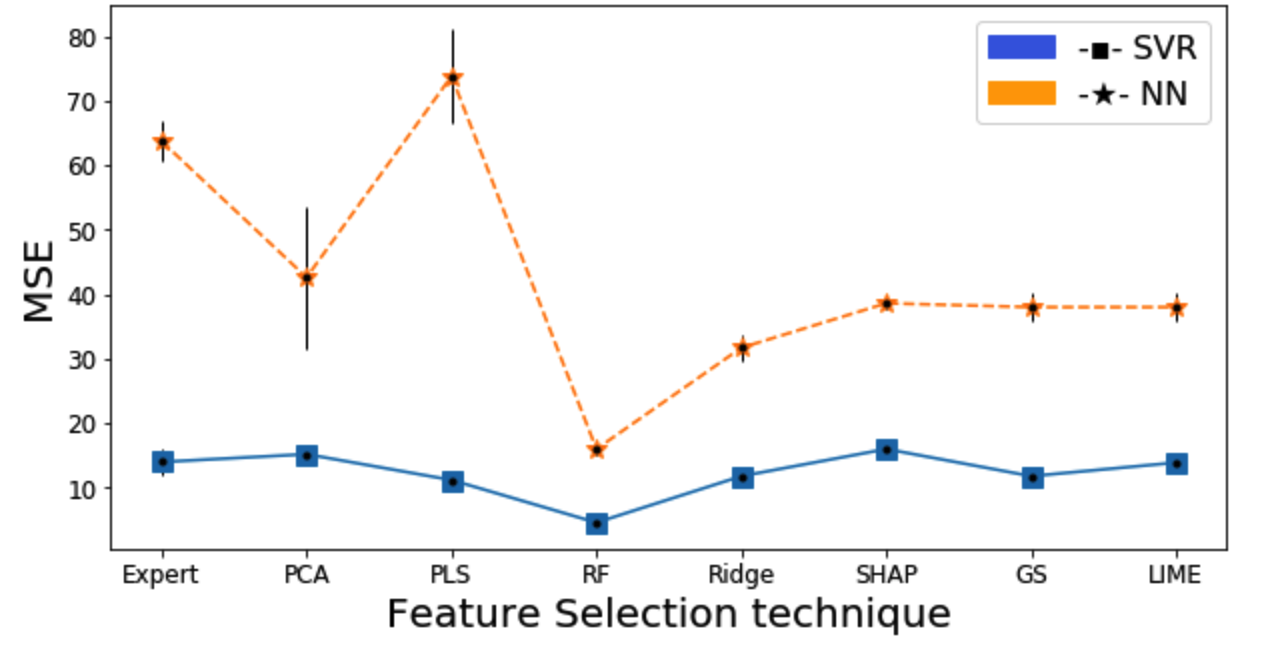}
    \caption{Testing error for Fine-Tuned SVR and NN models on Reduced subsets in Real-time scenario.}
    \label{fig:SVRvsNN}
\end{figure}

The ``curse of dimensionality''\cite{bellman2015adaptive} affects both models equally, therefore only through a careful selection of hyperparameters we are able to avoid excessive overfitting. Although we reach similar performance for NN in some cases compared to SVR, parameter optimization for SVR, in general, is a much easier task, as fewer parameters are considered. As a result, after thorough tuning, Fine-Tuned SVR performs significantly better across all three scenarios (Figure \ref{fig:SVRvsNN}), with LR performing the worst. Notably, the Reduced Fine-Tuned SVR and NN models trained on features selected by Random Forest have significantly lower testing error and variance in Real-time scenarios compared to the Full model. To that extent, the Fine-Tuned SVR model with RF features produces the lowest Test MSE out of all models for Real-time scenarios and is, therefore, selected as the best performing feature selection technique. We can explain the superior performance of RF feature selection in that it creates decision trees on boot-strapped subsets of observed data and randomly selected subsets of features (as previously discussed) before taking the majority vote on features that reduce the variance. Since this ensemble model considers different combinations of data and features, it has a better ability to identify important attributes for prediction. Since we are dealing with a small number of samples, one or two observations can really change overall data distribution. Therefore, RF has a better sense of variance when translated to unobserved data and results in better generalization ability.

Using the Fine-Tuned SVR model as a benchmark, we observe that Reduced models trained on features selected by Ridge, SHAP, GS, and LIME have competitive testing error with that of expert-selected features. One explanation, as mentioned earlier, is that not all expert features affect the CN prediction, while features selected with the above-mentioned techniques actually do. 

PCA and PLS show the most unexpected results since these revered methods perform noticeably worse under different settings in both correctness and performance categories compared to all other methods. One possible explanation for their previous widespread use in chemometric analysis is the assumption that most chemical processes and reactions are linear. Both PCA and PLS decompose the data into linear combinations of original features and are commonly paired with linear regression in the literature.
However, given the non-linearity of spectroscopic data, presence of noise, and overall high dimensionality, a further attempt at selecting a reduced number of important attributes results in poor performance of these methods. Further, PCA and PLS have simple and efficient implementation. Therefore, at first glance, complicated methods that can be superior to PCA and PLS were excluded from consideration in the literature until now.

\begin{figure}[!htbp]
\centering
\begin{subfigure}{\textwidth}
    \includegraphics[width=1\linewidth]{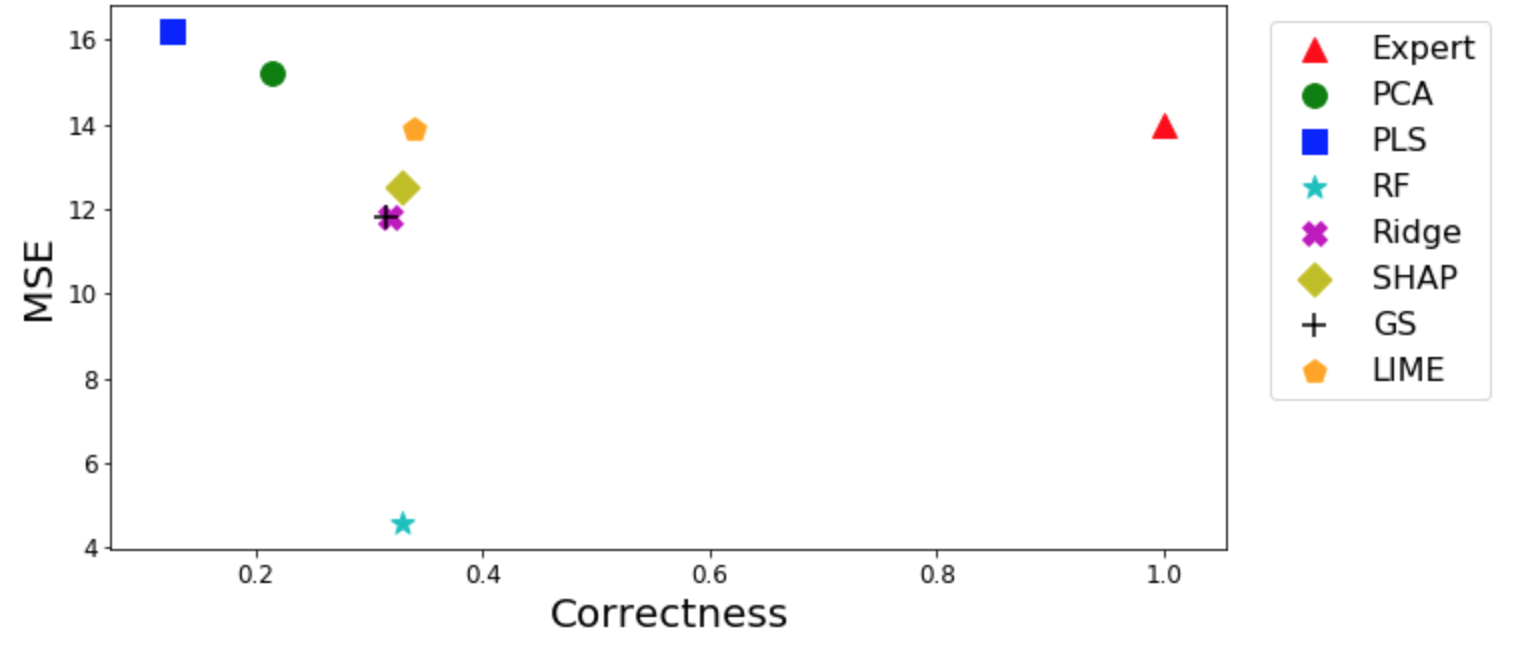}
    \caption{SVR}
    \label{fig:TradeoffA}
\end{subfigure}%

\begin{subfigure}{\textwidth}
    \includegraphics[width=1\linewidth]{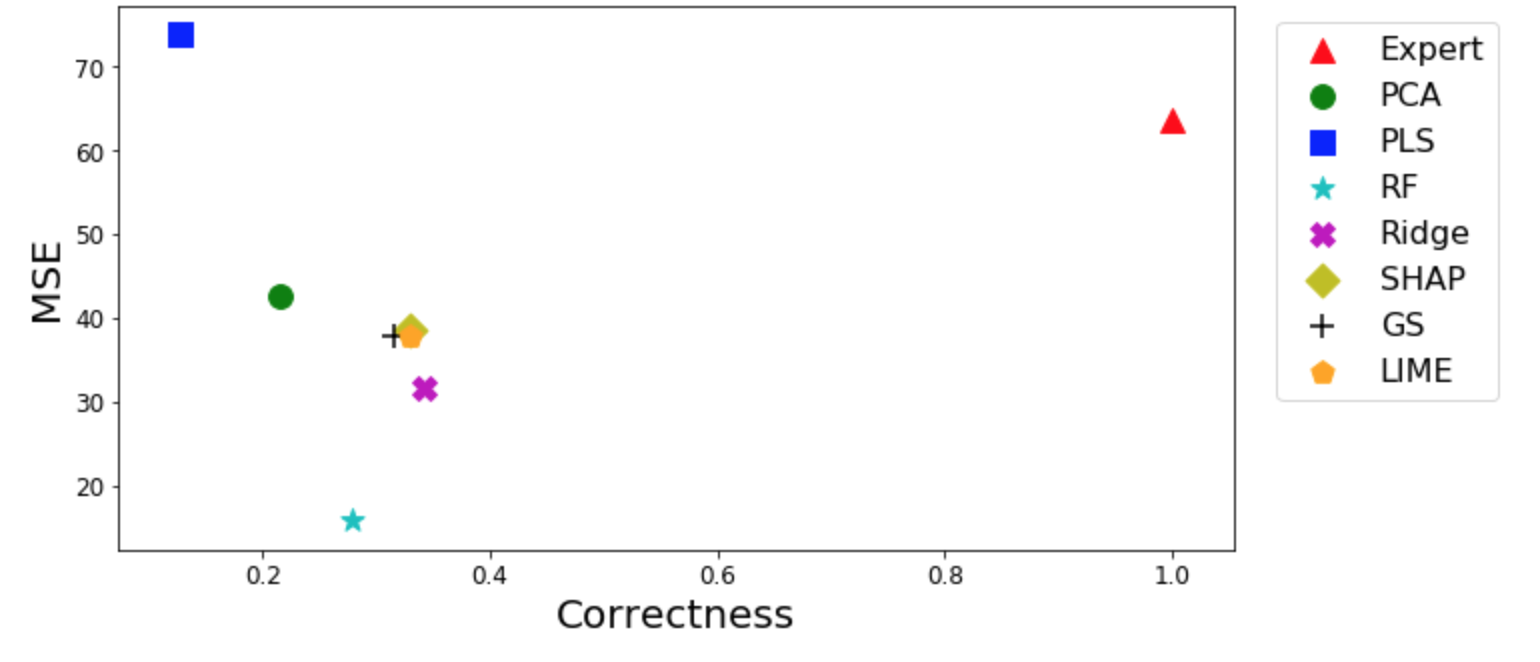}
    \caption{NN} 
    \label{fig:TradeoffB}
\end{subfigure}%

\caption{Testing Error on Real-time dataset vs Correctness trade-off for 120 selected features using (a) SVR and (b) NN models}
\label{fig:Tradeoff}
\end{figure}

Figure \ref{fig:Tradeoff} demonstrates the Correctness versus Performance trade-off for both Fine-Tuned SVR and NN models, we can conclude that Random Forest selects the most meaningful subsets of attributes both from an explainability and prediction performance standpoint. This paper shows that optimizing hyperparameters for NN can be a challenging task that requires paying significant attention to the underlying data distribution. We also determine that given the high-dimensional limited size of data, SVR predictive model outperforms NN, although a further increase in the size of data can tip the scales back in favor of NN.

\section{Conclusion}\label{sec:conclusion}

This paper identified a gap in the scalability and interpretability of ML-based spectroscopy data analysis literature. To address it, we applied a diverse and comprehensive set of ML tools to solve a real-world chemometrics problem and explain the machine decision process behind the solution to any domain practitioner. We provided a detailed description of the theoretical background and implementation considerations for all techniques used. Finally, we interpreted the ML results from both the domain and data science perspectives to promote congruent explanations.

Our experiment results demonstrated that applying less complex ML predictive models is preferred for fuel spectroscopy analysis over linear or more complex network-based techniques.
A small training sample size, inherent non-linearity in the underlying relationship of features with the response, the presence of the noise in the data, and 
processing of highly-dimensional data are major reasons for such observations. 

Constructing support vector decision boundary has shown to have better scaling and generalization power for real-world deployment and proved to yield better Cetane Number prediction results than Neural Network model. The choice of hyperparameters is crucial task to to handle overfitting for both models, the task which is more complicated in NN. 

Using various feature selection techniques, we were able to confirm that the reduced subset of features in general results in a less complex, computationally more efficient, and yet accurate prediction model. Based on our interpretability evaluation, we were also able to confirm the conformance of the attribute selection and prediction results in terms of known chemistry, ensuring that the domain expert can trust such algorithmic results. We were able to explain the behavior of complex ML models using various model-agnostic methods, which has not been done before in chemometrics to this extent, and report on performance of model-based methods. As discovered, using model-agnostic explanation techniques, features derived from complex ML models were generally of higher quality than from model-based methods, with the exception of subset of Random Forest features that performed best overall on Performance-Explainability trade-off scale. While the PCA and PLS decomposition are still considered as the primary and widely-used techniques for spectroscopy data analysis, our investigation concluded that these methods were inferior to all other feature extraction techniques covered in this work.

\section{Acknowledgement}

The authors thank Professor Ken Brezinsky, who facilitated the data collection process using equipment available at UIC High-Pressure Shock Tube Laboratory and advised them throughout the data collection process. Dr. Patrick Lynch has also advised the authors at different stages of this project, including data pre-processing and curation. This project is supported by the U.S. Army Research Laboratory grant W911NF2020223.

\bibliographystyle{elsarticle-num-names} 
\bibliography{ref}





\end{document}